\documentclass[sn-mathphys-num]{sn-jnl}

\usepackage{graphicx}%
\usepackage{multirow}%
\usepackage{amsmath,amssymb,amsfonts}%
\usepackage{amsthm}%
\usepackage{mathrsfs}%
\usepackage[title]{appendix}%
\usepackage{xcolor}%
\usepackage{textcomp}%
\usepackage{manyfoot}%
\usepackage{booktabs}%
\usepackage{algorithm}%
\usepackage{algorithmicx}%
\usepackage{algpseudocode}%
\usepackage{listings}%
\usepackage{makecell}%
\usepackage{ulem}
\usepackage{comment}
\usepackage{booktabs}
\usepackage{threeparttable}
\usepackage{tabularx}
\usepackage{float}

\theoremstyle{thmstyleone}%
\theoremstyle{thmstyletwo}%

\theoremstyle{thmstylethree}%

\begin{document}


\title[Measuring Deep Learning Performance - An Empirical Study of Performance Distributions Across Architectures and Tasks]{Measuring Deep Learning Performance - An Empirical Study of Performance Distributions Across Architectures and Tasks}


\author*[1,2]{\fnm{Kevin L} \sur{Coakley}}\email{kcoakley@sdsc.edu}

\author[1]{\fnm{Odd Erik} \sur{Gundersen}}\email{odderik@ntnu.no}

\affil*[1]{\orgdiv{Department of Computer Science}, \orgname{Norwegian University of Science and Technology}, \orgaddress{\city{Trondheim}, \country{Norway}}}

\affil[2]{\orgdiv{San Diego Supercomputer Center}, \orgname{University of California San Diego}, \orgaddress{\city{La Jolla}, \state{CA}, \country{USA}}}




\abstract{
\textbf{Context:} Non-determinism in deep learning algorithm design and implementation leads to performance variation, meaning model performance is not a single value, but rather a distribution. 
These model performance distributions are underexplored despite their impact on robustness.

\textbf{Objective:} We investigate the robustness of deep learning performance to sources of non-determinism, specifically focusing on how performance distributions differ across various architectures and tasks.

\textbf{Method:} We conducted 186 experiments on state-of-the-art image classification (ResNet, ViT) and time series forecasting (Autoformer, iTransformer, NLinear, TSMixer) architectures. 
Each experiment was run 100 times with different random seeds to generate performance distributions, resulting in 18,600 runs. Robustness was quantified using metrics for spread, symmetry, and tail risk.

\textbf{Results:} Performance distributions are frequently non-Gaussian, particularly in time series forecasting. 
Model size does not systematically affect robustness -- larger image classification models show fewer outliers but not lower spread, while smaller time series models show lower spread but more extreme underperformers. 
Training duration does not scale linearly; early stopping effectively balances performance and robustness. 
Mean performance does not predict robustness -- time series forecasting shows moderate correlation while image classification shows none. 
Time series models produce nearly three times more underperforming outliers than image classification models, indicating substantially higher tail risk.

\textbf{Conclusions:} Tail risk poses serious concerns for Trustworthy AI in high-stakes applications. 
Models performing well on average may exhibit long tails and extreme outliers revealed only through distributional analysis. 
Mean performance alone should not guide model selection; assessment of spread, symmetry, and tail risk is essential for reliable model assessment where consistent performance is critical.}






\maketitle

\section{Introduction}
\label{sec:introduction}

Many factors affect the performance of a deep learning model. 
Some are related to the way data are treated, such as how the data is divided into train, validation, and test sets \cite{berrar_2024,renard_2020}, while others are related to how the algorithm is designed, such as how hyperparameters are optimized \cite{bouthillier2021accounting,bouthillier_2019,renard_2020} and weights are initialized \cite{pham_2020,zhuang_2022}. 
The way an algorithm is implemented can also affect the performance, including the software package used to implement the algorithm \cite{henderson_2018,pouchard_2020,renard_2020}, the operating system and compiler settings used \cite{hong_2013}, and which seed is used to initialize the pseudo-random number generator \cite{melis2018state,gundersen_2023}.
The computational budget used to search for the optimal hyperparameters of a deep learning algorithm can also affect performance, since searching longer costs more but will increase the chance of finding a better set of hyperparameters \cite{dodge_2019,lucic_2018,zhang_2020}. 
These sources of variation in the performance cause problems for deep learning benchmarks both in regard to conducting fair comparisons and in evaluating the results \cite{raff_2020,gundersen_2022,gundersen_2023}.

As factors related to data and algorithm design depend on more or less active decisions, they are more easily controlled. 
Implementation factors, on the other hand, can be harder to control, as implementing the same algorithm using different software packages or running it on different hardware or operating systems with different compiler settings requires not only more effort, but also access to different computer systems.
The exception is the initialization of the pseudo-random number generator. 
The initialization seed can be easily controlled and, when controlled, can enable deterministic experiments -- with some exceptions.
Since the performance of a deep learning algorithm depends on the sequence of random numbers sampled during training and inference, the performance is not a single fixed value, but rather a distribution \cite{reimers_2017,gundersen_2023}.  
Despite this being a well-documented fact, model performance distributions are underexplored in the literature. 

A better understanding of model performance distributions is needed. 
The shape of model performance distributions and how they vary across different model architectures is not well understood. 
Although previous work establishes that model performance distributions are not Gaussian \cite{reimers_2017,pham_2020,gundersen_2023,turan_2026}, it remains unclear whether similar deep learning architectures exhibit similar performance distributions. 
Are some architectures naturally more resilient to non-determinism, while others are more prone to variability?
Are fat tails and skewed performance distributions issues?
What about extreme outliers?
Model performance distributions that are skewed, have fat tails, and extreme outliers cannot be considered robust. 
As robustness is a dimension of trustworthy artificial intelligence \cite{li2023trustworthy,kaur2022trustworthy,liang2022advances,chatila2021trustworthy}, model distributions that are skewed, have fat tails or extreme outliers might not be considered trustworthy 
and would pose a potential risk in high-stakes domains involving human safety or financial risk.

Deep learning has been applied across a range of domains where failures carry direct consequences for human welfare or financial stability. 
In medical imaging, deep learning models are used for tumor detection, disease classification, and organ segmentation from MRI, CT, and X-ray data\cite{jiang_2023,renard_2020,tran_2021}. 
In financial markets, time series forecasting models have been applied to asset price prediction and stock movement classification, including spatio-temporal graph neural network approaches for cryptocurrency and foreign exchange markets\cite{uygun_2025} and hypergraph neural network approaches that model higher-order relationships among stocks\cite{alaygut_2025}. 
In energy markets, deep learning models are used for electricity price forecasting, power production forecasts and load prediction, tasks that directly affect grid stability and market operations\cite{el_2024,manfre_2023,dalal2021day,foley2012current,ahmed2020review}. 
These are all high-stakes domains where incorrect or inconsistent model predictions carry real-world consequences, making it essential to investigate not only mean performance, but the full distribution of outcomes under non-deterministic training.

Therefore, in this paper, we investigate the robustness of deep learning performance with respect to sources of non-determinism. 
As all sources of variability have similar effects on model diversity\cite{summers_2021} and are non-additive \cite{zhuang_2022}, we limit our study to controlling the initialization seed of pseudo-random number generators.
This allows us to investigate non-determinism by running the same experiment many times and only varying one of its sources.
This allows us to run the same experiment 100 times while ensuring such subtle differences and therefore getting a better representation of the underlying true distribution of the model performance.
As we present the results of a much larger number of experiments than other studies of variation have presented, we provide more reliable evidence. 
In contrast to most previous studies, we do not limit the experiments to imaging tasks but also include time series forecasting algorithms to compare results between tasks. 
Our main contributions are as follows:
\begin{itemize}
    \item We present the first study on the performance distributions of deep learning models with a focus on robustness.
    \item We empirically establish that the best performing model is not necessarily the most robust one. 
    \item We empirically establish that deep learning time series forecasting models are less robust than deep learning image classification models.  
\end{itemize}

The rest of the article is structured as follows. 
First, in Section \ref{sec:related-work}, we present related work, and then in Section \ref{sec:problem-description}, we describe the problem we investigated in detail along with the research questions we seek to answer.
We describe the experiments in Section \ref{sec:experimental-design} and provide a detailed description of how we measure the robustness of model performance distributions in Section \ref{sec:evaluation-metrics}.
Then, in Section \ref{sec:results}, the results are presented, and we discuss all of our findings in Section \ref{sec:discussion}. 
Finally, in Section \ref{sec:conclusion}, we conclude. 

\section{Related work}
\label{sec:related-work}
As the inherent randomness of a deep learning model leads to variations in the result of the inference, its performance can be considered a variable with uncertain outcome.  
\citet{hullermeier2021aleatoric} distinguish between two types of uncertainty, namely aleatoric and epistemic uncertainty. 
Aleatoric uncertainty is uncertainty due to inherently random effects, such as the result of a coin flip, while epistemic uncertainty is caused by a lack of knowledge and, as such, could be reduced by improving the model. 
The model performance can be interpreted as aleatoric uncertainty, as it is caused by inherently random effects in the training and reasoning process.
This is, at least in effect, how it has been treated, as we have found literature that observes this uncertainty but none that sought to mitigate it. 
Only recently, with the increased focus on reproducibility, has this uncertainty been commonly accepted, which is reflected by the introduction of reproducibility checklists for the main machine learning conferences; see \citet{pineau2021improving}. 
They emphasize that not only central tendencies of the results should be reported, but also their variation, which was not the standard before the checklists were introduced. 

Mitigating the performance uncertainty would be a way to increase the robustness of the inference process.
\citet{dietterich_2017} proposes that AI methods applied in high-stakes settings must be robust to both \textit{known unknowns} and \textit{unknown unknowns}.
Known unknowns are the uncertain aspects of the world that the model can reason explicitly with, while unknown unknowns are those aspects that are not captured by the model. 
The reason he focuses on high-stakes settings is, of course, that robustness is more important in such settings where failures have high costs, whether it is measured in monetary value, environmental damages, or human lives.
Dietterich proposes four ideas to improve robustness to known unknowns, which are robust optimization, regularization in machine learning, risk-sensitive objectives, and robust inference. 
Ideas for how to handle unknown unknowns include detecting model failures, using causal models, relying on a portfolio of models, and expanding the model.

In this article, however, we are interested in something that has been less investigated and understood, namely the robustness of the inherent randomness of a model due to randomness in its training and implementation. 
In light of \citet{dietterich_2017}, we consider the inherent stochasticity of a model as a known unknown, as it is known that deep learning models have this characteristic. 
However, as far as we know, this randomness has only been observed, and solutions to reduce it have not been proposed in the literature, although there is no reason it could not be done if modeled explicitly.

\subsection{The Sources of Randomness}
\label{subsec3}

\citet{pham_2020} categorized factors that cause variability into algorithmic and implementation-level sources of non-determinism. 
Algorithmic factors are methods that use randomization to enhance training efficiency and model accuracy, such as nondeterministic layers (e.g. dropout), random weight initialization, data augmentation, and stochastic batch ordering. 
\citet{pham_2020} found that these algorithmic factors could cause up to 10.8\% variation in model accuracy, even after eliminating weaker models.
Implementation factors, on the other hand, come from the hardware and software used to train machine learning models. These include parallel computation processes, auto-selection of primitive operations, and floating-point rounding errors. 
\citet{pham_2020} demonstrated that implementation factors alone could lead to variations of up to 2.9\% in overall accuracy and 52.4\% in per-class accuracy.

In their study, \citet{summers_2021} examined various sources of non-determinism in deep learning optimization, identifying algorithmic and implementation factors (random initialization of parameters, data shuffling, and low-level library operations) as significant contributors to run-to-run variability. 
Their findings reveal that even minimal perturbations, such as altering a single weight by the smallest possible amount within machine precision, can lead to substantial differences in model performance. 
This variability is observed across different model architectures and depths, highlighting the sensitivity of deep learning models to seemingly minor perturbations.
Finally, they note that all sources of non-determinism have similar effects on the results of deep learning model training.

Zhuang et al. \cite{zhuang_2022} further investigated the effects of algorithmic and implementation randomness on machine learning training, reinforcing and extending the findings of \citet{pham_2020}. 
They concluded that addressing non-determinism within a single component of the technical stack is ineffective for decreasing overall training variability. 
Instead, they assert that non-determinism must be comprehensively managed across all levels of the technical stack or not at all, noting that interactions between different sources of non-determinism are non-additive.

\subsection{Impact of Variability Caused by Non-Determinism on Machine Learning Model Performance}

The inherent non-determinism present in the training of deep learning models is a well-acknowledged characteristic that gives rise to variability in performance outcomes. 
Even with identical model architectures and datasets, repeated training runs can yield a range of results. 
Accurately characterizing this variability is essential for building trustworthy AI systems that can be relied upon for critical tasks. 
This section reviews prior research on the impact of non-deterministic variation in reported model performance, highlighting the challenges of drawing reliable conclusions from single training instances.

The paper “Unreproducible Research is Reproducible” by Bouthillier et al.\cite{bouthillier_2019} investigates how the current practice of benchmarking deep learning models for image classification tasks is sensitive to algorithmic and implementation sources of non-determinism, leading to unreliable conclusions about model performance. 
The authors demonstrate through experiments that “concluding which model performs best based on observations from a single initialization seed is brittle: this conclusion will often be falsified if using a different seed. This is especially true for simpler datasets, but one also sees that model ranking varies widely across datasets.” 
They argue that to ensure the robustness and inferential reproducibility of conclusions, it is crucial to account for sources of variation like random initialization and data sampling in the experimental design, highlighting the need for the machine learning community to adopt more rigorous empirical methodologies.
A singular representation of deep learning model results does not capture the full picture of a model's performance, which is in reality a distribution rather than a fixed value.

Studies have highlighted the importance of considering this distribution when evaluating machine learning model robustness. 
Reimers and Gurevych\cite{reimers_2017} demonstrated the impact of random weight initialization via random seeds, where different seeds can lead to statistically significant variations in performance metrics. 
They examined two state-of-the-art systems for named entity recognition, which reported F1-scores of 90.94\% and 91.21\%. 
However, when these implementations were rerun multiple times with different random seeds, the model with the lower reported F1-score exhibited superior performance when considering the distribution quartiles. 

Gundersen et al.\cite{gundersen_2023} further illustrated this point by examining the N-BEATS and DeepAR models on the exchange\_rate dataset. 
Their analysis revealed that the performance variation within a single model due to different random seeds was greater than the variation between the two models. 
This suggests that random seed selection can have a more substantial impact on model performance than the choice between these specific models.

These studies collectively underscore the importance of considering both algorithmic and implementation sources of non-determinism when evaluating model robustness and reproducibility in machine learning systems. 
They caution that the variability introduced by these factors can significantly influence the perceived superiority of one model over another, highlighting the need for more comprehensive evaluation methods that account for this inherent variability.

\subsection{Machine Learning Model Performance Distributions}

\citet{turan_2026} conducted a systematic investigation of generalization performance distributions along learning curves for classical classifiers. 
Using 1000 repetitions per training set size across 10 datasets, they found that performance distributions rarely follow Gaussian distributions regardless of dataset balance, loss function, or hyperparameter tuning. 
They demonstrated that model rankings can change significantly when using different statistical measures, with a 94\% probability of different top-3 model rankings when comparing the $\text{mean} \pm 2\sigma$ versus the 0.975-quantile. 
While their work focused on classical machine learning classifiers with varying training set sampling and sizes, our study examines deep learning architectures with fixed datasets but multiple random seeds, extending the analysis across different tasks.

\subsection{Robustness to Tail Performance}\label{subsec2}

\citet{dietterich_2017} outlines various strategies to enhance the robustness of AI systems, particularly in high-stakes applications where failures can result in significant negative consequences. 
The paper covers several approaches, including robust optimization, regularization in machine learning, risk-sensitive objectives, robust inference, model failure detection, causal models, portfolio methods, and model expansion.

The section on Risk-Sensitive Objectives focuses on developing decision-making policies in Markov Decision Processes that emphasizes downside risks instead of only maximizing expected rewards.
Dietterich discusses using Conditional Value at Risk (CVaR) as an alternative to traditional expected total reward optimization.
CVaR measures the mean value of outcomes within a specific worst-case segment of a probability distribution as determined by a quantile parameter $\alpha$, typically 0.05.
CVaR optimization shifts the focus to the poorest-performing segment of potential outcomes, aiming to maximize the expected value within this specific subset. 
This approach allows for more conservative decision-making in scenarios where downside risk is a significant concern. 

Dietterich also highlights regularization as an important idea for achieving robustness in machine learning systems. 
Regularization techniques are widely adopted to prevent overfitting by constraining model complexity, thereby improving how well models generalize to new data. 
Significantly, Dietterich points out that regularization can be reformulated as “finding a robust optimum against an adversary capable of perturbing the data points.” 
This interpretation means that regularization seeks solutions that are not only effective on the training data but are also inherently stable, ensuring their predictions remain consistent despite minor input variations.

\section{Methodology}
\label{sec:methodology}

\subsection{Problem Description}
\label{sec:problem-description}

Deep learning model performance exhibits inherent variability that stems from various sources of non-determinism. 
While this variability has been acknowledged in the related work (Section \ref{sec:related-work}), its systematic characterization across different architectures and application tasks remains largely unexplored. 
Our work addresses this gap by investigating how different deep learning architectures respond to seemingly inconsequential sources of randomness in their training, an understanding critical for building more trustworthy AI systems.

The conventional approach of representing model performance as a single metric (such as accuracy or error rate) does not characterize the true nature of deep learning outcomes.
When trained with different random initializations but otherwise identical configurations, deep learning models produce a spectrum of performance results, a distribution rather than a fixed value. 
This distribution comes from the complex interaction between the algorithmic and implementation factors used during the training processes.

We define robustness of a machine learning model’s performance as the absence, or low level, of variation in the face of noise introduced by algorithmic and implementation factors. 
The opposite of robustness is sensitivity, which corresponds to large variation in performance of the machine learning model in the face of perturbation introduced by these factors.
A perfectly robust model will always produce the same output and thus perform exactly the same regardless of the algorithmic and implementation factors. 
In contrast, the performance of a highly sensitive model will vary widely if small perturbations are brought about by implementation and algorithmic factors.

This notion of robustness is not merely a technical aspiration but a critical component of building trustworthy AI systems. 
In fact, the European Commission's High-Level Expert Group on AI\cite{eu_2019} identifies technical robustness, alongside lawfulness and ethical behavior, as one of the three foundational pillars of Trustworthy AI, essential for minimizing unintentional harm.

Robustness represents a critical but often overlooked property when evaluating models, especially in the context of developing Trustworthy AI.
A highly sensitive model may occasionally achieve impressive peak performance but could also produce catastrophically poor results in certain training runs. 
Such unpredictability is particularly problematic in high-stakes domains like medical diagnostics, autonomous driving, or financial modeling, where reliability and consistent performance is essential.

Current practices in evaluating deep learning robustness suffer from two key limitations:

\begin{itemize}
    \item \textbf{Limited sampling of the performance distribution:} Most studies rely on small samples (typically three-ten runs with different random seeds) to characterize model performance. Such limited sampling provides an incomplete and potentially misleading picture of the true performance distribution, particularly for assessing tail risks.
    \item \textbf{Narrow focus on single application task:} Investigations of model robustness typically focus on a single task, making it difficult to determine whether findings about robustness generalize across different tasks or are specific to particular data structures and learning problems.
\end{itemize}

Our study addresses these limitations through two key methodological improvements:

\begin{itemize}
    \item \textbf{Comprehensive sampling:} We evaluate each model architecture using one hundred different random seeds, providing a much more detailed characterization of the performance distribution than is typical in the literature. This approach enables more reliable estimation of distribution parameters and better assessment of tail risks.
    \item \textbf{Cross-task comparison:} We systematically compare robustness across two fundamentally different tasks: image classification and time series forecasting. These tasks differ substantially in their data structures (spatial/grid-structured versus temporal/sequential) and modeling approaches, allowing us to identify both task-specific and task-general patterns in robustness.
\end{itemize}

To guide our investigation, we formulated the following set of research questions. 
These questions are designed to systematically examine not only the statistical properties of deep learning performance distributions, but also how robustness is affected by model size, training duration, performance, and task. 

\begin{itemize}
\item \textbf{RQ1:} Are deep learning model performance distributions Gaussian?
\item \textbf{RQ2:} Does the size of deep learning models affect their robustness?
\item \textbf{RQ3:} Does increasing the number of training epochs affect the robustness of deep learning models?
\item \textbf{RQ4:} Does robustness correlate with performance?
\item \textbf{RQ5:} Are image classification models more robust than time series forecasting models?
\end{itemize}

By addressing these questions, we aim to provide a more comprehensive understanding of robustness in deep learning model performance distributions.

\subsection{Experimental Design}
\label{sec:experimental-design}

We conducted controlled experiments to investigate the robustness of deep learning model performance distributions to non-determinism across two distinct tasks: image classification and time series forecasting. 
Our experiments evaluated multiple architectures, including Convolutional Neural Networks (CNNs), Vision Transformers (ViTs), Multi-Layer Perceptrons (MLPs), and linear models.

\textbf{Image Classification.} For image classification, we evaluated ResNet CNNs\cite{he_2016} (depths: 18, 20, 50, 56, 101, 110) and Vision Transformers in Tiny\cite{steiner_2021} (ViT-T), Small\cite{steiner_2021} (ViT-S), and Base\cite{dosovitskiy_2020} (ViT-B) configurations. 
We used four datasets: CIFAR-10 and CIFAR-100\cite{krizhevsky_09} (32×32 images) for smaller-scale evaluation, and Oxford Flowers\cite{nilsback_08} and UC Merced\cite{yang_2010} (256×256 images) for larger-scale assessment. 
This combination enabled comprehensive analysis across 100 training runs per model-dataset pair while maintaining computational feasibility. 
ResNet and ViT models were trained using random weight initialization, the ViT models were also trained with pretrained weights from HuggingFace's Timm library\footnote{\href{https://huggingface.co/docs/timm/}{https://huggingface.co/docs/timm/}}.

\textbf{Time Series Forecasting.} For time series forecasting, we examined NLinear\cite{zeng_2023} (linear), TSMixer\cite{chen_2023} (MLP), and Autoformer\cite{wu_2021} and iTransformer\cite{liu_2023} (transformers) using six datasets: four Electricity Transformer Temperature (ETT) datasets\cite{zhou_2021}, Traffic\cite{wu_2021}, and Weather\cite{wu_2021}. 
We adapted the code shared by the original authors, with modifications to ensure consistent sequence (96) and horizon lengths (96, 192, 336, 720) and deterministic training. 
Traffic horizons 336 and 720 were not completed for iTransformer due to GPU memory constraints on the NVIDIA A100s.

\textbf{Experimental Framework.} To systematically investigate the impact of algorithmic and implementation factors on model robustness, we adopt three core concepts defined by \citet{bouthillier_2019}:

A \textit{model architecture} is a specific parameterized function form together with its standard recommended random parameter initialization strategy. 
This encompasses the complete structure and design of a neural network, including its layers, connections, and hyperparameter configuration.
The model architecture defines what remains constant across repeated training runs.

A \textit{seed replicate} is a single instance of a model architecture trained and evaluated with a specific random seed value. 
This random seed determines the model's initial parameters and the order of training data presentation. 
Each seed replicate represents one sample from the model's performance distribution and can only be exactly reproduced using deterministic training with the identical seed.

A \textit{dataset replicate} is a set of seed replicates for a given model architecture on a specific dataset. 
A performance distribution can be created from the dataset replicate for that model-dataset combination under non-deterministic training conditions.

For each model-dataset combination, we trained 100 independent seed replicates with identical hyperparameters and data splits, varying only the random initialization seed. 
We maintained experimental fidelity to the original implementations, adopting published configurations, training procedures, and evaluation protocols to ensure that observed performance distributions reflect non-determinism rather than methodological divergence.
This close adherence to established experimental setups ensures that our observed model performance distributions genuinely reflect the impact of non-determinism rather than methodological divergences.

Our design isolates seed-driven variation within a fixed software and hardware environment. 
The resulting performance distributions therefore characterize random-seed variability under controlled implementation conditions, rather than the joint ALGO+IMPL setting described by \citet{zhuang_2022}. 
Since algorithmic and implementation sources of non-determinism interact non-additively\cite{zhuang_2022}, our results should not be interpreted as a direct test of that interaction.

Full details of the deterministic configuration for both frameworks, including PyTorch and TensorFlow environment settings, are provided in Appendix \ref{sec:appendix-training}.

Experiments were executed on an HPC cluster\cite{sjalander_2019} with NVIDIA A100 GPUs using containerized environments to ensure reproducibility. 
Deterministic behavior was verified across compute nodes by replicating single-epoch training with identical seeds. 
This design enabled robust characterization of performance distributions: we conducted 186 experiments (28 image classification, 158 time series forecasting), each comprising 100 training runs, totaling 18,600 runs. 
Complete implementation details are provided in Appendix \ref{sec:apx-experiment-details}.

\subsection{Evaluation Metrics}
\label{sec:evaluation-metrics}

\subsubsection{Performance Metrics}

We employ task-specific metrics to evaluate model performance across different domains.

\textbf{Top-1 Accuracy} measures the percentage of test samples for which the model's highest-confidence prediction matches the true class label in image classification tasks. 
Higher values indicate better performance. 
For cross-task comparisons with error-based metrics, we use top-1 error (1 - accuracy) as the complement, allowing direct comparison with regression-based measures.

\textbf{Mean Absolute Error (MAE)} quantifies forecasting accuracy in time series tasks by computing the average magnitude of prediction errors. 
While both MAE and Mean Squared Error (MSE) appear commonly in forecasting literature, we exclusively use MAE because MSE's squared term disproportionately penalizes large errors, potentially skewing cross-task robustness comparisons.

\subsubsection{Robustness Metrics}
\label{sec:robustness-metrics}

We quantify robustness along three statistical dimensions of model performance distributions: spread, symmetry, and tail risk.

\textbf{Spread} characterizes overall variability across seed replicates. We measure spread using \textbf{range} and \textbf{standard deviation}. 
Lower values indicate greater robustness. 
Range captures the total span of outcomes but is sensitive to extreme values, while standard deviation incorporates all data points and provides a more stable measure of dispersion.

\textbf{Symmetry} describes how performance values distribute around the central tendency. 
\textbf{Skewness} quantifies distributional asymmetry: positive values indicate longer right tails, negative values indicate longer left tails. 
The interpretation depends on the metric: for accuracy-based measures, negative skew suggests increased likelihood of underperforming replicates; for error-based measures, positive skew indicates similar risk. 
Values closer to zero reflect more symmetric, robust distributions. 
In skewed distributions, mean performance may not represent typical outcomes, as the mean can be pulled toward the tail.

\textbf{Tail Risk} captures the frequency of extreme outcomes using \textbf{outlier counts} based on Tukey's fences (1.5 × IQR). 
While we initially considered CVaR, as discussed by \citet{dietterich_2017} for risk-sensitive optimization, CVaR and its related metric Value-at-Risk (VaR) are fundamentally quantile-based measures -- VaR corresponds to a specific quantile (typically the 5th percentile) of the loss distribution, while CVaR averages values beyond that threshold. 
Similar quantile-based approaches have been employed by \citet{turan_2026} for characterizing performance distributions along learning curves. 
However, these metrics are sensitive to shifts in central tendency and most effective when distributions share similar scales and centers. 
This makes them unsuitable for comparing model performance distributions across tasks that use different performance metrics. 
Instead, we employ Tukey's fences, a non-parametric approach that operates independently of performance metrics and scales, making it suitable for cross-task comparison. 
We define the interquartile range as IQR = $Q_3$ - $Q_1$, where $Q_1$ and $Q_3$ are the 25th and 75th percentiles. 
\textbf{Lower fence outliers} fall below $Q_1$ - 1.5 × IQR, representing unusually poor performance. 
\textbf{Upper fence outliers} exceed $Q_3$ + 1.5 × IQR, indicating exceptionally strong performance. 
For accuracy-based metrics, lower fence outliers indicate worse performance; for error-based metrics, upper fence outliers indicate worse performance.

\subsubsection{Statistical Testing}

We apply standard statistical tests to evaluate research questions systematically. 
The \textbf{Shapiro–Wilk test} ($\alpha = 0.05$) assesses normality of performance distributions (RQ1). 
The \textbf{binomial test} ($p = 0.5$) determines whether binary outcomes deviate from chance expectations (RQ1 and RQ2).
To supplement the binomial test for RQ2, we applied the paired \textbf{Wilcoxon signed-rank test} on raw robustness metric values, with rank-biserial correlation as an effect size measure.
\textbf{Pearson correlation coefficient} measures linear relationships between continuous variables, such as training epochs and robustness (RQ3 and RQ4). 
For RQ4, we additionally report \textbf{Spearman's} $\boldsymbol{\rho}$ to assess the monotonic relationship between performance and robustness without assuming linearity.
For comparing means between independent samples, we use \textbf{Welch's t-test} when variances differ and \textbf{Student's t-test} when variances are similar (RQ5). 
To account for the number of comparisons across research questions, we applied the \textbf{Holm-Bonferroni correction} to each family of tests, where each research question constitutes a separate family.
Each test was selected based on its suitability for the inference type and relevance to robustness analysis.

\section{Results}
\label{sec:results}

\subsection{Are Deep Learning Model Performance Distributions Gaussian?}
\label{sec:are-mpd-gaussian}

\begin{figure}[!t]
    \centering
    \includegraphics[width=1\linewidth]{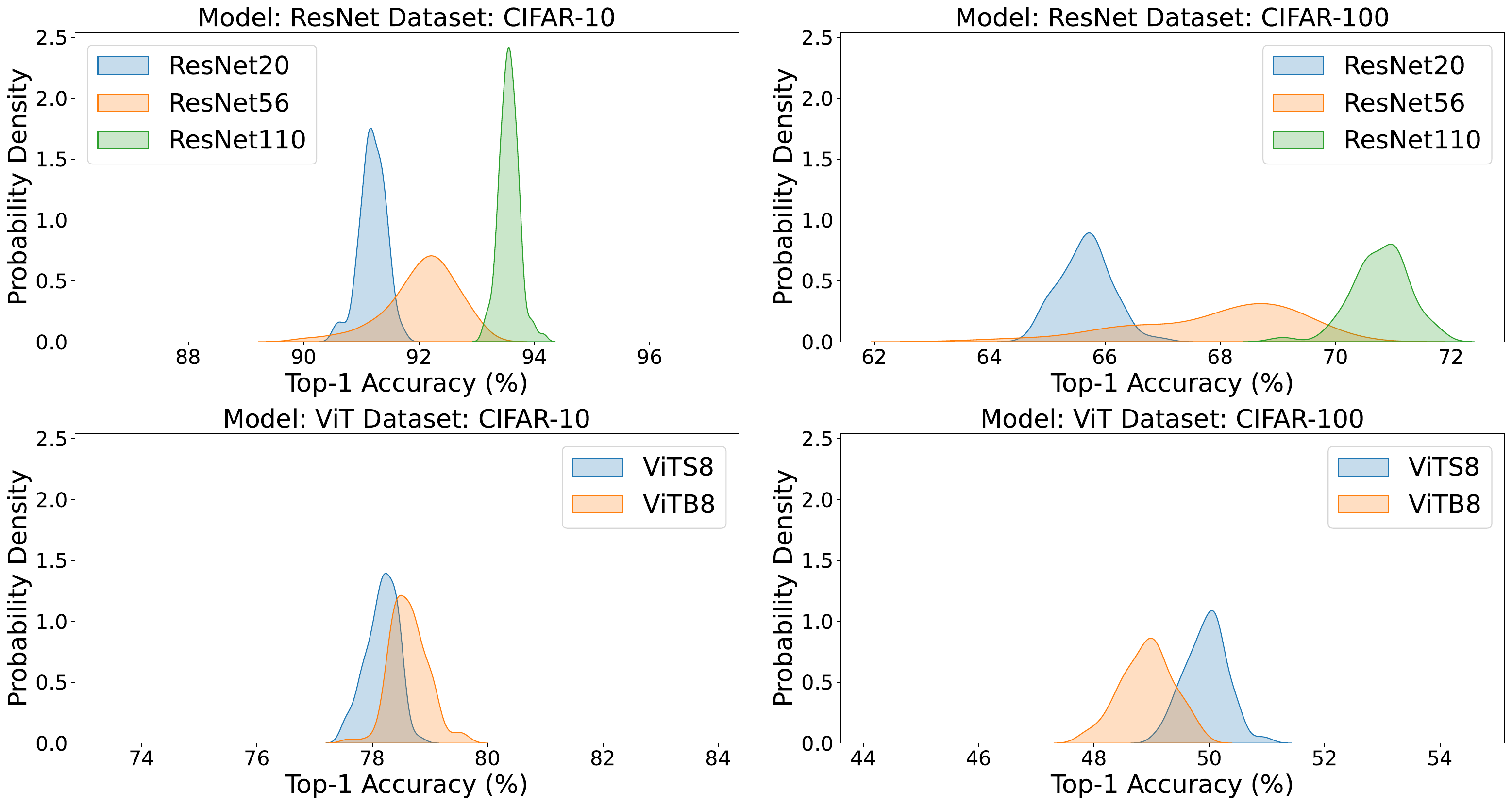}
    \caption{Kernel density estimates of top-1 accuracy for image classification models on CIFAR-10 (left) and CIFAR-100 (right). Top row shows ResNet architectures (ResNet20, ResNet56, ResNet110); bottom row shows Vision Transformers (ViT-S-8, ViT-B-8). Each curve represents the distribution across 100 random seeds for a single model-dataset pair.}
    \label{figure:results_cifar_kde}
\end{figure}

\begin{figure}[!t]
    \centering
    \includegraphics[width=1\linewidth]{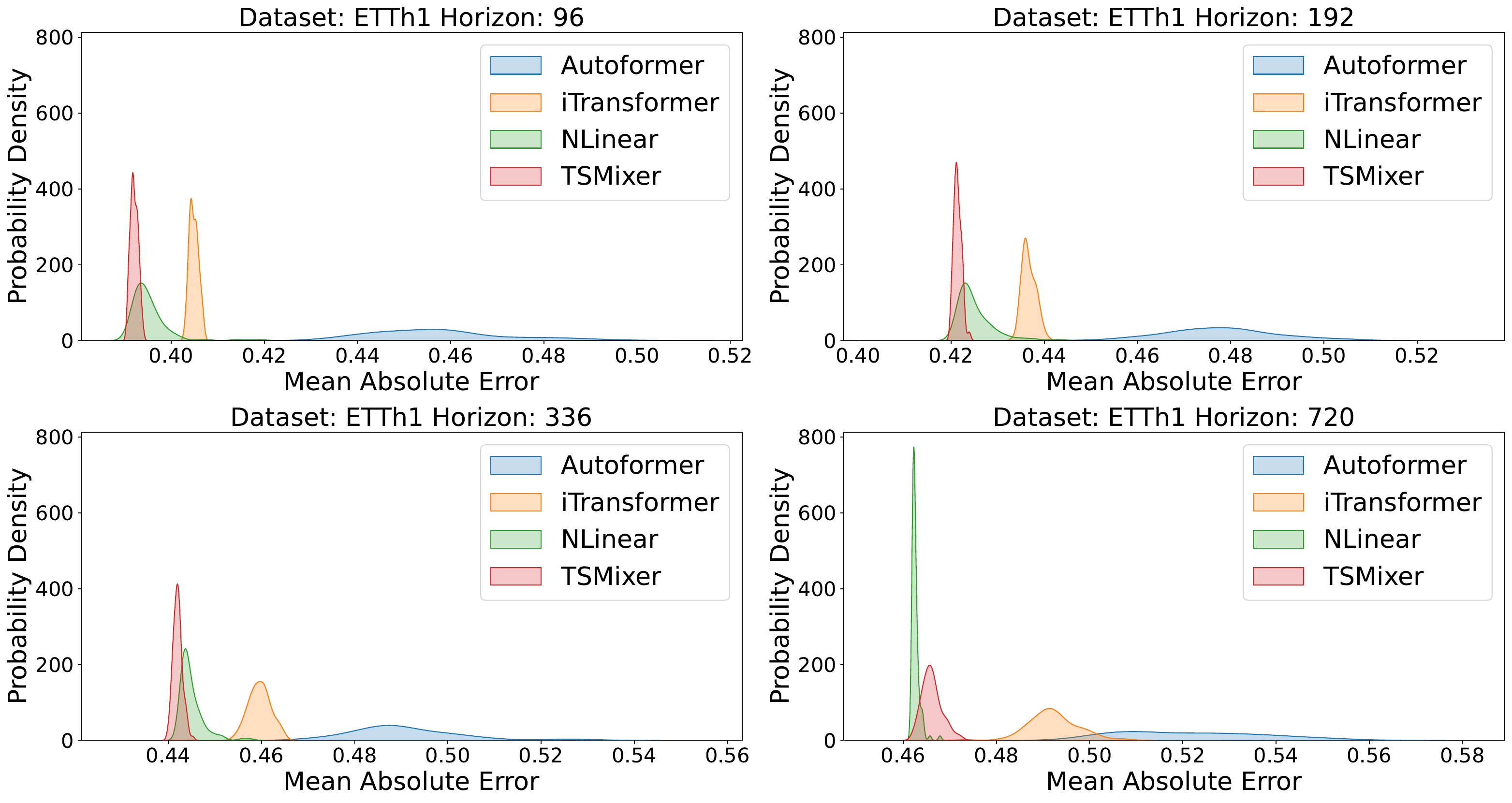}
    \caption{Kernel density estimates of MAE for time series forecasting models on ETTh1 across forecast horizons of 96, 192, 336, and 720. Each panel shows distributions for Autoformer, iTransformer, NLinear, and TSMixer. Each curve represents the distribution across 100 random seeds for a single model-horizon pair.}

    \label{figure:results_etth1_kde}
\end{figure}

Prior literature\cite{pham_2020,gundersen_2023,turan_2026} suggests that machine learning and deep learning model performance distributions are not Gaussian and recommends using non-parametric statistical tests for model comparison. 
We investigate this empirically across a large set of experiments.

We analyzed 122 model performance distributions: 28 from image classification and 94 from time series forecasting. 
Performance was measured using top-1 accuracy for image classification and MAE for time series forecasting. 
Each distribution comprises 100 seed replicates trained with identical configurations but different random initialization seeds.

To determine whether each distribution was Gaussian, we applied the Shapiro–Wilk test with a significance threshold of $\alpha = 0.05$. 
Distributions with $p < 0.5$ were classified as non-Gaussian.
Because the p-value alone is highly sensitive at $n = 100$, where the test has sufficient power to reject normality for trivially small deviations, we report the W statistic alongside the p-value as an effect-size measure for non-normality.
The Shapiro–Wilk W statistic ranges from 0 to 1, indicating the proportion of variance explained by a normal distribution. 
Values of W $\geq$ 0.98 indicate the distribution is approximately normal regardless of a significant p-value, values between 0.95 and 0.98 indicate mild deviations, and W $<$ 0.95 indicates a substantively non-Gaussian distribution.

Results are reported in \autoref{tab:ic_results} (image classification) and Tables \ref{tab:ts_mae_results_1}, \ref{tab:ts_mae_results_2}, and \ref{tab:ts_mae_results_3} (time series forecasting). 
Kernel density estimates of the performance distributions are shown in Figures \ref{figure:results_cifar_kde} and \ref{figure:results_etth1_kde} for representative datasets (all figures are in \autoref{sec:additional_figures}). 
We then conducted a binomial test under the null hypothesis that Gaussian and non-Gaussian outcomes are equally likely ($p = 0.5$) to assess whether non-Gaussian behavior is systematic across dataset replicates.

Of the 122 distributions evaluated, 71 were classified as non-Gaussian: 12 out of 28 in image classification and 59 out of 94 in time series forecasting.
However, because the Shapiro-Wilk test is highly sensitive at n=100, we evaluated the practical magnitude of these deviations using the W statistic as an effect-size measure. Among the 71 distributions with a p-value $<$ 0.05, none exhibited W $\geq$ 0.98, 32 showed mild deviations (0.95 $\leq$ W $<$ 0.98), and 39 demonstrated substantive non-Gaussian shapes (W $<$ 0.95). Notably, the vast majority of these substantive deviations (37 out of 39) occurred in time series forecasting tasks. In contrast, non-Gaussian image classification distributions largely exhibited only mild deviations (10 out of 12).

The binomial test indicated that non-Gaussian distributions were significantly more common than expected by chance across all distributions (p = 0.042), driven by time series forecasting ($p = 0.009$), while image classification showed no significant deviation from chance ($p = 0.828$). 
After Holm-Bonferroni correction, the time series forecasting result remains significant (Holm-adjusted $p = 0.027$) while the overall result does not (Holm-adjusted $p = 0.084$). 
Therefore, after correction, the evidence for systematic non-Gaussian behavior is specific to time series forecasting and does not extend to the overall distribution of results.

The visualizations in Figures \ref{figure:results_cifar_kde} and \ref{figure:results_etth1_kde} illustrate this pattern. 
Image classification models generally exhibit relatively symmetric distributions, though with some variation in spread. 
Time series forecasting models, however, show more pronounced deviations from normality, including skewness, particularly visible in \autoref{figure:results_etth1_kde} for the ETTh1 dataset.
Deep learning model performance distributions are frequently non-Gaussian, particularly in time series forecasting. 
Our findings align with prior work and support the use of non-parametric statistical methods for model evaluation, even when performance distributions are derived from substantially more seed replicates than in prior studies.

\subsection{Does the Size of Deep Learning Models Affect their Robustness?}
\label{sec:does-size-affect-robustness}

\begin{table}[!t]
\centering
\resizebox{\textwidth}{!}{%
\footnotesize
\setlength{\tabcolsep}{4pt}
\begin{tabular}{lccccccc}
\toprule
\textbf{Robustness} & \multicolumn{2}{c}{\underline{\textbf{\# of Models More Robust}}} & \textbf{Equally} & \textbf{\# of} & \multicolumn{2}{c}{\underline{\textbf{p-value}}} \\
\textbf{Measure} & \textbf{Larger Models} & \textbf{Smaller Models} & \textbf{Robust} & \textbf{Trials} & \textbf{Larger Models} & \textbf{Smaller Models} \\
\midrule
\multicolumn{7}{l}{\textit{Intra-architecture Binomial Comparison}} \\
\midrule
Range & 12 & 14 & 0 & 26 & 0.7214 & 0.4225 \\
Standard Deviation & 9 & 17 & 0 & 26 & 0.9622 & 0.0843 \\
Skewness & 14 & 12 & 0 & 26 & 0.4225 & 0.7214 \\
Lower Fence Outliers & 12 & 10 & 4 & 22 & 0.4159 & 0.7383 \\
Upper Fence Outliers & 10 & 7 & 9 & 17 & 0.3145 & 0.8338 \\
All Outliers & 14 & 6 & 6 & 20 & 0.0577 & 0.9793 \\
\midrule
\multicolumn{7}{l}{\textit{Cross-architecture Binomial Comparison}} \\
\midrule
Range & 28 & 27 & 1 & 55 & 0.5 & 0.6061 \\
Standard Deviation & 22 & 34 & 0 & 56 & 0.9593 & 0.0704 \\
Skewness & 36 & 20 & 0 & 56 & \textbf{0.022} & 0.9889 \\
Lower Fence Outliers & 29 & 16 & 11 & 45 & \textbf{0.0362} & 0.9822 \\
Upper Fence Outliers & 21 & 17 & 18 & 38 & 0.3136 & 0.7912 \\
All Outliers & 30 & 12 & 14 & 42 & \textbf{0.004} & 0.9986 \\
\bottomrule
\end{tabular}
} 
\caption{Binomial test results for robustness comparisons between image classification models with different parameter counts. Rows show counts of pairwise comparisons favoring larger models, smaller models, or ties, with corresponding p-values. Hypothesized probability $p = 0.5$ for all tests.}
\label{tab:ic_binomial}
\end{table}

\begin{table}[!t]
\centering
\resizebox{\textwidth}{!}{%
\footnotesize
\setlength{\tabcolsep}{4pt}
\begin{tabular}{lccccccc}
\toprule
\textbf{Robustness} & \multicolumn{2}{c}{\underline{\textbf{\# of Models More Robust}}} & \textbf{Equally} & \textbf{\# of} & \multicolumn{2}{c}{\underline{\textbf{p-value}}} \\
\textbf{Measure} & \textbf{Larger Models} & \textbf{Smaller Models} & \textbf{Robust} & \textbf{Trials} & \textbf{Larger Models} & \textbf{Smaller Models} \\
\midrule
\multicolumn{7}{l}{\textit{Cross-architecture Binomial Comparison}} \\ 
\midrule
Range & 27 & 111 & 0 & 138 & 1.0 & $\boldsymbol{1.36 \times 10^{-13}}$ \\
Standard Deviation & 27 & 111 & 0 & 138 & 1.0 & $\boldsymbol{1.36 \times 10^{-13}}$ \\
Skewness & 83 & 55 & 0 & 138 & \textbf{0.0106} & 0.9934 \\
Upper Fence Outliers & 78 & 47 & 13 & 125 & \textbf{0.0035} & 0.998 \\
\bottomrule
\end{tabular}
} 
\caption{Binomial test results for robustness comparisons between time series forecasting models with different parameter counts. Rows show counts of pairwise comparisons favoring larger models, smaller models, or ties, with corresponding p-values. Hypothesized probability $p = 0.5$ for all tests.}
\label{tab:ts_binomial}
\end{table}

We examine whether model size, measured by the number of parameters, affects the robustness of deep learning performance distributions. 
We analyzed results from 122 dataset replicates: 28 for image classification and 94 for time series forecasting. 
Robustness was quantified using spread (range and standard deviation), symmetry (skewness), and tail risk (counts of lower and upper fence outliers), as defined in Section \ref{sec:robustness-metrics}.

For image classification, we conducted both intra-architecture (e.g., ResNet20 vs. ResNet56) and cross-architecture (e.g., ResNet vs. ViT) comparisons. 
Comparisons were only made between models trained using the same weight initialization strategy; models trained with random initialization were not compared to those trained with pretrained weights. 
For time series forecasting, only cross-architecture comparisons were performed, limited to models trained on the same dataset and horizon length. 
Full robustness results are reported in \autoref{tab:ic_results} (image classification) and Tables \ref{tab:ts_mae_results_1}, \ref{tab:ts_mae_results_2}, and \ref{tab:ts_mae_results_3} (time series forecasting). 
Binomial test outcomes are summarized in Tables \ref{tab:ic_binomial} (image classification) and \ref{tab:ts_binomial} (time series forecasting).

To assess whether model size systematically affects robustness, we performed pairwise comparisons of robustness metrics across dataset replicates. 
For each comparison, we identified two models trained on the same dataset, where one model had more parameters than the other. 
We then compared their robustness to determine which model was more robust.

A model was considered more robust if it had lower values for spread metrics (range or standard deviation), skewness closer to zero (lower absolute skew) for symmetry, or fewer outliers for tail risk. 
We aggregated pairwise outcomes and used the binomial test to determine whether larger or smaller models were more robust more often than expected by chance. 
The null hypothesis was that both outcomes were equally likely ($p = 0.5$). 
Each test was conducted in both directions. 
When results were equal, we excluded those trials from the binomial tests.
Due to the high number of equal results, comparisons involving lower fence outliers and total outlier counts were excluded from the time series analysis.

\subsubsection{Image Classification}

For intra-architecture comparisons, the binomial test revealed no statistically significant differences for any robustness metric. 
For spread, larger models showed smaller ranges in nearly equal numbers (12 versus 14), while smaller models had lower standard deviations approximately twice as often (17 versus 9). 
For symmetry, larger models had less skew in nearly equal instances (14 versus 12). 
For tail risk, larger models had fewer lower fence outliers in nearly equal instances (12 versus 10). 
Upper fence outliers were relatively balanced (10, 7, and 9 for larger fewer, smaller fewer, and equal, respectively). 
Total outliers favored larger models approximately twice as often (14 versus 6).

For cross-architecture comparisons, spread metrics showed no statistically significant differences. 
Larger models showed smaller ranges in nearly equal instances (28 versus 27), while smaller models had lower standard deviations more often (34 versus 22).

For symmetry, the binomial test yielded a nominally significant result for skewness ($p = 0.022$), though this does not survive Holm-Bonferroni correction.
Larger models had less skew than smaller models (36 versus 20).

For tail risk, the binomial test indicated nominally significant differences for lower fence outliers ($p = 0.0362$) and all outliers ($p = 0.004$), neither of which survives Holm-Bonferroni correction.
Larger models had fewer lower fence outliers nearly twice as often (29 versus 16). 
Upper fence outliers were relatively balanced (21, 17, and 18). 
Total outliers favored larger models approximately twice as often (30 versus 12).

Our pairwise comparisons reveal that larger models tend to have fewer outliers but do not consistently exhibit smaller range or standard deviation. 
The lack of statistical significance in intra-architecture comparisons, despite smaller models exhibiting more outliers, is likely due to limited pairwise trials. 
Cross-architecture comparisons, with more trials and the same directional trend, yielded sufficient power to detect statistically significant effects.

Larger models appear to produce more uniformly spread performance distributions, with results dispersed more evenly across the interquartile range, leading to higher standard deviations despite fewer extreme values. 
Smaller models show more concentrated distributions, with results clustered more tightly around the center but with higher outlier frequency. 
Cross-architecture results indicate that larger models have more Gaussian (less skewed) distributions and fewer lower fence and overall outliers.

\subsubsection{Time Series Forecasting}

For spread metrics, the binomial test revealed statistically significant differences for both range and standard deviation ($p = 1.36 \times 10^{-13}$ for both; Holm-adjusted $p = 2.18 \times 10^{-12}$)
Smaller models showed smaller ranges and standard deviations more than four times as often (111 versus 27).

For symmetry, the binomial test yielded a nominally significant result for skewness ($p = 0.0106$), though this does not survive Holm-Bonferroni correction.
Larger models had less skew than smaller models (83 versus 55).

For tail risk, the binomial test indicated statistically significant differences for upper fence outliers ($p = 0.0035$; Holm-adjusted $p = 0.049$).
Larger models had fewer upper fence outliers (78 versus 47).

Our analysis reveals distinct patterns in time series forecasting. 
Smaller models consistently demonstrated lower spread, while larger models exhibited more symmetric distributions and fewer underperformance outliers. 
These patterns were most pronounced at the extremes of the model size spectrum. 
The largest model, Autoformer, consistently showed the widest range and standard deviation across nearly all datasets and forecast horizons, while the smallest model, NLinear, exhibited the greatest skewness and highest number of lower fence outliers.

When examining only intermediate-sized models (iTransformer and TSMixer), results for range, standard deviation, and upper fence outliers were nearly balanced, with only skewness showing a non-significant preference toward larger models. 
These findings suggest that model-specific behaviors, rather than model size alone, may substantially contribute to the observed robustness patterns.

\subsubsection{Summary}

In image classification, larger models exhibited more robust performance distributions with respect to symmetry and outlier counts, particularly in cross-architecture comparisons. 
However, larger models did not consistently outperform smaller models in spread metrics, indicating that increased size does not uniformly lead to more robust performance distributions.

In time series forecasting, smaller models consistently demonstrated lower spread, while larger models showed more symmetric distributions and fewer extreme underperformers. 
However, these effects were largely driven by the most extreme models (Autoformer and NLinear). 
Analysis of intermediate-sized models showed more balanced results.

In both tasks, model size influences some aspects of robustness but not others. 
Because these effects are inconsistent across robustness metrics and differ between tasks, we conclude that model size does not systematically affect robustness.

To supplement the binomial test results, we computed paired Wilcoxon signed-rank tests and rank-biserial correlation coefficients on the raw robustness metric values for each pairwise model size comparison (Tables~\ref{tab:ic_wilcoxon} and~\ref{tab:ts_wilcoxon}). 
After applying Holm-Bonferroni correction, the Wilcoxon results are largely consistent with the binomial findings. 
For image classification, no comparisons survive correction under either framework, and the rank-biserial correlations confirm that effect sizes are small to negligible throughout. 
For time series forecasting, range, standard deviation, and upper fence outliers survive correction under both frameworks, with large effect sizes for range ($r = 0.758$) and standard deviation ($r = 0.805$) confirming that the spread advantage of smaller models is practically meaningful. 
The one discrepancy between the two frameworks occurs for skewness: the binomial result does not survive correction, whereas the Wilcoxon result does (Holm-adjusted $p = 0.0091$, $r = -0.332$), indicating a medium effect favoring larger models in distributional symmetry.

After Holm-Bonferroni correction, only the time series spread results and upper fence outliers remain significant, reinforcing the conclusion that model size does not systematically affect robustness.

\subsection{Does Increasing the Number of Training Epochs Affect the Robustness of Deep Learning Models?}
\label{sec:increasing-epochs-affect-robustness}

\begin{figure}[!t]
    \centering
    \includegraphics[width=1\linewidth]{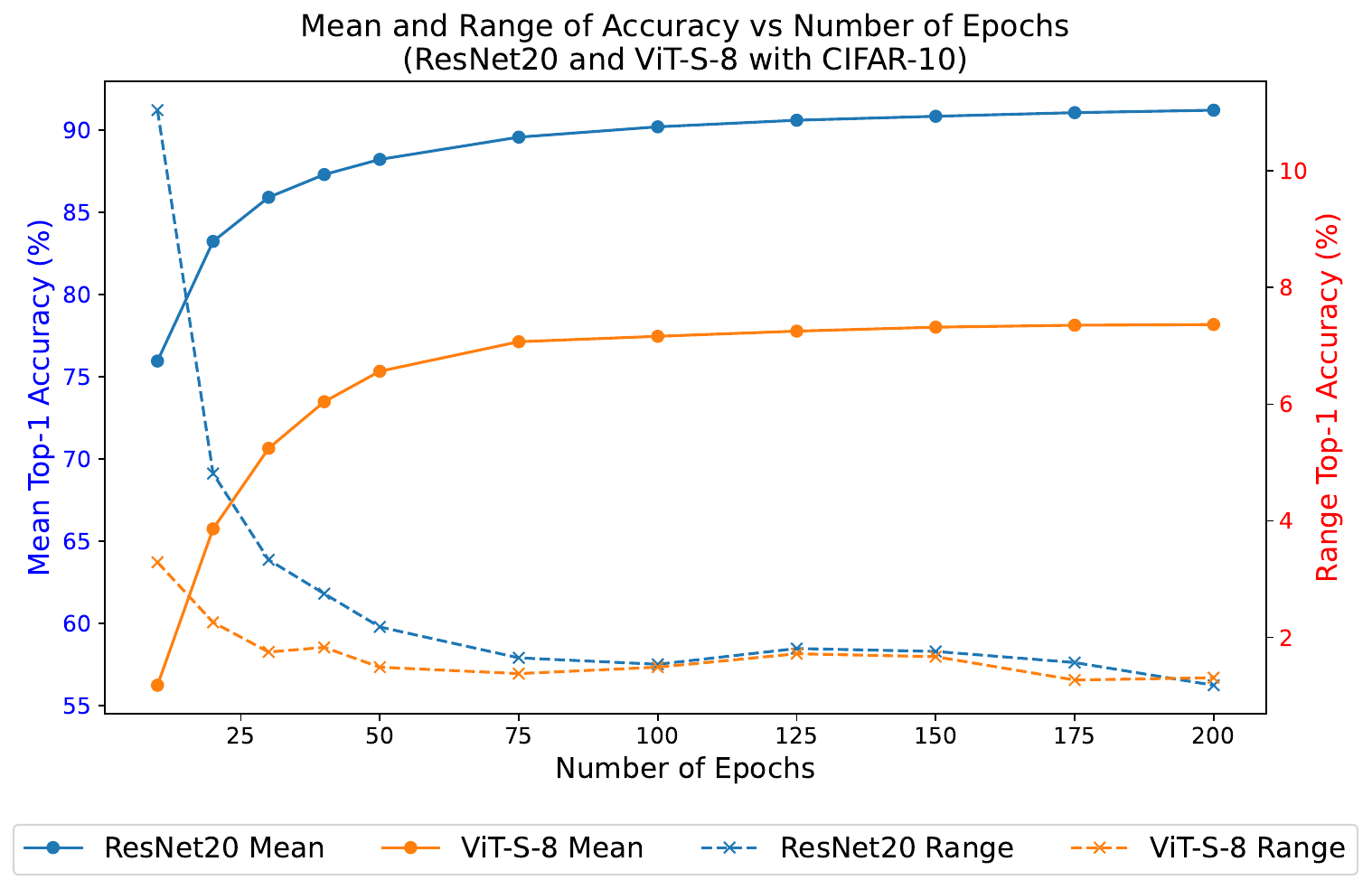}
    \caption{Mean and range of top-1 accuracy across training epochs for ResNet20 (blue) and ViT-S-8 (orange) on CIFAR-10. Solid lines show mean accuracy (left y-axis); dashed lines show range (right y-axis). Each point computed from 100 seed replicates.}
    \label{fig:epochs_image}
\end{figure}

\begin{figure}[!t]
    \centering
    \includegraphics[width=1\linewidth]{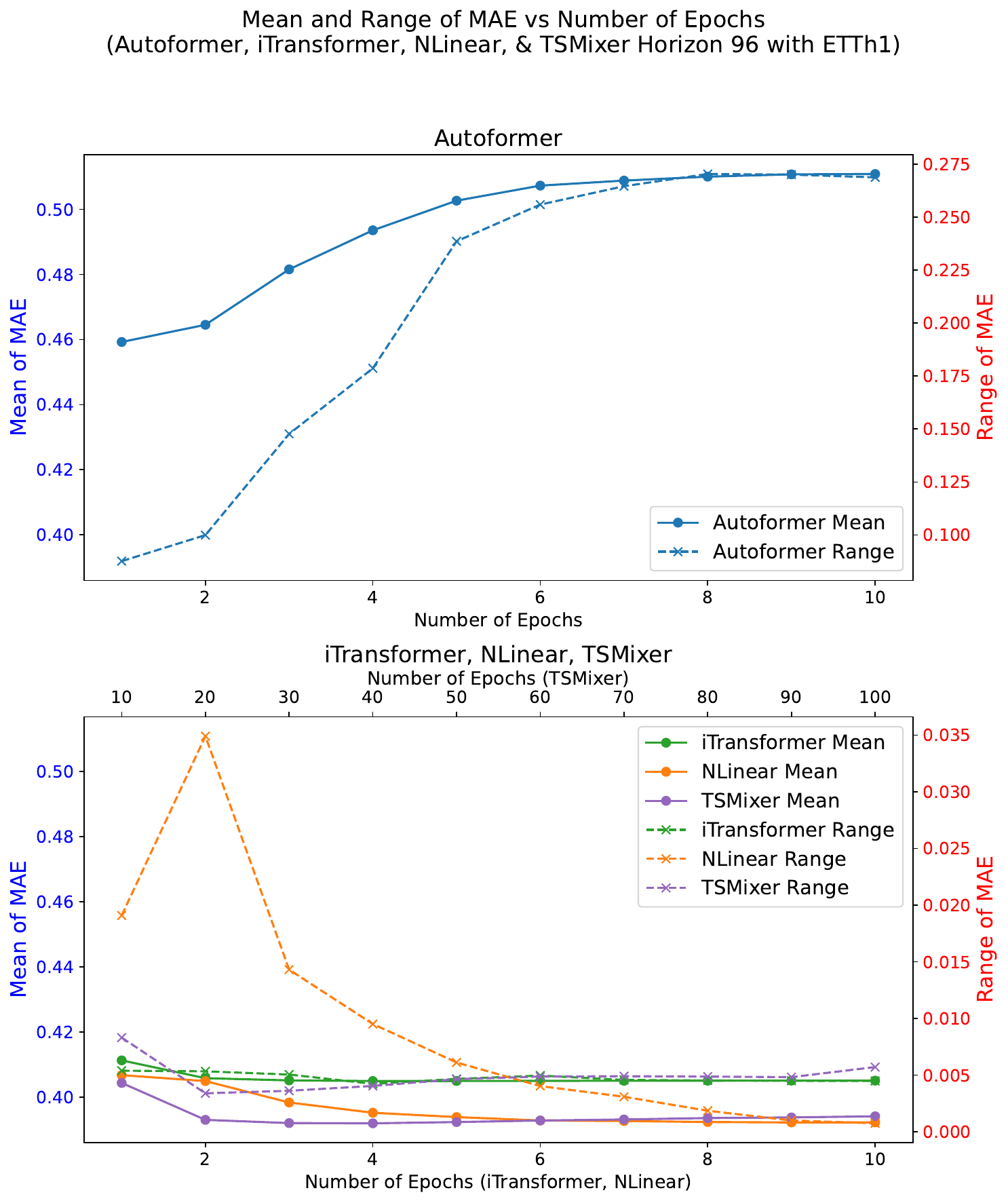}
    \caption{Mean MAE (solid lines, left y-axis) and range (dashed lines, right y-axis) across training epochs for time series forecasting models on ETTh1 with forecast horizon 96. Top panel shows Autoformer (epochs 1-10); bottom panel shows iTransformer and NLinear (epochs 1-10, bottom x-axis) and TSMixer (epochs 10-100, top x-axis). Each point computed from 100 seed replicates trained for the specified number of epochs.}
    \label{fig:epochs_time_series}
\end{figure}

\begin{table}[h!]
  \centering
  \label{tab:ic_epochs}
  \begin{tabular}{ccccccc}
    \toprule
    \multicolumn{3}{c}{ResNet20} & \multicolumn{3}{c}{ViT-S-8} \\
    \cmidrule(lr){1-3} \cmidrule(lr){4-6}
    Epochs & Mean & Range & Epochs & Mean & Range \\
    \midrule
    10 & 75.94\% & 11.04\% & 10 & 56.23\% & 3.29\% \\
    20 & 83.22\% & 4.81\% & 20 & 65.74\% & 2.26\% \\
    30 & 85.90\% & 3.33\% & 30 & 70.64\% & 1.75\% \\
    40 & 87.28\% & 2.75\% & 40 & 73.47\% & 1.83\% \\
    50 & 88.20\% & 2.18\% & 50 & 75.32\% & 1.49\% \\
    75 & 89.56\% & 1.65\% & 75 & 77.12\% & 1.38\% \\
    100 & 90.18\% & 1.54\% & 100 & 77.45\% & 1.49\% \\
    125 & 90.58\% & 1.81\% & 125 & 77.76\% & 1.72\% \\
    150 & 90.82\% & 1.76\% & 150 & 78.00\% & 1.67\% \\
    175 & 91.04\% & 1.57\% & 175 & 78.12\% & 1.27\% \\
    200 & 91.19\% & 1.18\% & 200 & 78.16\% & 1.31\% \\
    \midrule
    \multicolumn{3}{c}{Correlation: $-0.9752$} & \multicolumn{3}{c}{Correlation: $-0.9528$} \\
    \bottomrule
  \end{tabular}
\caption{Mean top-1 accuracy and range for ResNet20 and ViT-S-8 on CIFAR-10 across training epochs. Each row computed from 100 seed replicates. Pearson correlation coefficients show the relationship between epochs and range (robustness).}
\end{table}

\begin{table}[h!]
  \centering
  \resizebox{\textwidth}{!}{%
  \label{tab:ts_epochs}
  \begin{tabular}{ccccccccccccc}
    \toprule
    \multicolumn{3}{c}{Autoformer} & \multicolumn{3}{c}{iTransformer} & \multicolumn{3}{c}{NLinear} & \multicolumn{3}{c}{TSMixer} \\
    \cmidrule(lr){1-3} \cmidrule(lr){4-6} \cmidrule(lr){7-9} \cmidrule(lr){10-12}
    Epochs & Mean & Range & Epochs & Mean & Range & Epochs & Mean & Range & Epochs & Mean & Range \\
    \midrule
    1 & 0.4593 & 0.0876 & 1 & 0.4112 & 0.0054 & 1 & 0.4067 & 0.0191 & 10 & 0.4043 & 0.0083 \\
    2 & 0.4645 & 0.0999 & 2 & 0.4058 & 0.0053 & 2 & 0.4049 & 0.0349 & 20 & 0.3930 & 0.0034 \\
    3 & 0.4816 & 0.1477 & 3 & 0.4051 & 0.0051 & 3 & 0.3983 & 0.0143 & 30 & 0.3920 & 0.0036 \\
    4 & 0.4936 & 0.1787 & 4 & 0.4049 & 0.0042 & 4 & 0.3951 & 0.0095 & 40 & 0.3919 & 0.0041 \\
    5 & 0.5027 & 0.2387 & 5 & 0.4049 & 0.0047 & 5 & 0.3938 & 0.0061 & 50 & 0.3923 & 0.0046 \\
    6 & 0.5073 & 0.2560 & 6 & 0.4049 & 0.0050 & 6 & 0.3928 & 0.0040 & 60 & 0.3928 & 0.0049 \\
    7 & 0.5089 & 0.2647 & 7 & 0.4050 & 0.0046 & 7 & 0.3926 & 0.0031 & 70 & 0.3931 & 0.0049 \\
    8 & 0.5101 & 0.2704 & 8 & 0.4050 & 0.0045 & 8 & 0.3923 & 0.0019 & 80 & 0.3935 & 0.0049 \\
    9 & 0.5108 & 0.2701 & 9 & 0.4050 & 0.0045 & 9 & 0.3922 & 0.0010 & 90 & 0.3937 & 0.0048 \\
    10 & 0.5109 & 0.2688 & 10 & 0.4050 & 0.0045 & 10 & 0.3922 & 0.0008 & 100 & 0.3941 & 0.0057 \\
    \midrule
    \multicolumn{3}{c}{Correlation: $0.9902$} & \multicolumn{3}{c}{Correlation: $0.6204$} & \multicolumn{3}{c}{Correlation: $0.9041$} & \multicolumn{3}{c}{Correlation: $0.9165$} \\
    \bottomrule
  \end{tabular}
  } 
\caption{Mean MAE and range for Autoformer, iTransformer, NLinear, and TSMixer on ETTh1 across training epochs. Each row computed from 100 seed replicates. Pearson correlation coefficients show the relationship between epochs and range (robustness).}
\end{table}

We investigate whether increasing the number of training epochs influences the robustness of deep learning model performance distributions by saving model performance at multiple points throughout training.

For image classification, we evaluated ResNet20 and ViT-S-8 on CIFAR-10, training each for 11 different epoch counts (10, 20, 30, 40, 50, 75, 100, 125, 150, 175, and 200) with 100 seed replicates per epoch. 
For time series forecasting, we evaluated Autoformer, iTransformer, NLinear, and TSMixer on ETTh1. Models were trained for up to 10 epochs (Autoformer, iTransformer, NLinear) or 100 epochs in 10-epoch intervals (TSMixer). 
Early stopping was disabled to ensure full training completion.

Robustness was measured using the range of the model performance distribution across 100 seed replicates for each epoch. Performance was measured as top-1 accuracy for image classification and MAE for time series forecasting. Results are reported in Tables \ref{tab:ic_epochs} and \ref{tab:ts_epochs} and visualized in Figures \ref{fig:epochs_image} and \ref{fig:epochs_time_series}.

To assess whether increasing epochs affects robustness, we computed the Pearson correlation coefficient between mean performance and range. 
For image classification, a negative correlation indicates that robustness improves (range decreases) as accuracy increases. 
For time series forecasting, a positive correlation indicates robustness improves (range decreases) as MAE decreases. 
We visualized the joint evolution of performance and robustness using dual-axis plots (Figures \ref{fig:epochs_image} and \ref{fig:epochs_time_series}).

For both ResNet20 and ViT-S-8 on CIFAR-10, mean accuracy improved steadily with additional epochs (Figure 5). 
However, robustness did not consistently improve. ResNet20's range decreased steadily to 100 epochs (1.54\%) but then fluctuated at epochs 125, 150, and 175 before decreasing again at 200 epochs (1.18\%). 
The correlation between accuracy and range was strongly negative ($r = -0.9752$). 
ViT-S-8 showed fluctuating robustness, with ranges varying between 1.31\% and 1.72\%, though the correlation was also strongly negative ($r = -0.9528$).

For time series models, increasing epochs did not consistently improve performance (Figure \ref{fig:epochs_time_series}). 
NLinear showed consistent improvement in both MAE and range across most epochs. 
In contrast, Autoformer exhibited increasing MAE and range, with a strong positive correlation ($r = 0.9902$), indicating worsening robustness as performance declined. 
iTransformer and TSMixer exhibited their best robustness early (epochs 4 and 2, respectively), after which both MAE and range increased. 
The correlations were positive for both models ($r = 0.6204$ for iTransformer, $r = 0.9165$ for TSMixer).

While longer training generally improved performance, its effect on robustness was inconsistent. 
For image classification, robustness did not consistently improve despite accuracy gains. 
For time series forecasting, robustness often peaked early and degraded with further training, except for NLinear. 
However, when considering early stopping, the epoch with best performance typically also exhibited strong robustness. 
For all time series models, the lowest MAE and lowest or near-lowest range occurred at the same epoch: epoch 1 for Autoformer, epoch 4 for iTransformer, epoch 9 for NLinear, and epoch 40 for TSMixer.

Although early stopping was disabled to observe the full training trajectory, the epoch-level results approximate its effect: for each model, the epoch with the best mean performance corresponds to the point at which early stopping would have triggered, and robustness at that epoch was at or near its strongest. 
We note that this approximation assumes early stopping triggers at the same epoch across all seeds; in practice, different seeds may converge at different rates, producing a distribution where each seed contributes performance from a different epoch.

\subsection{Does Robustness Correlate with Performance?}
\label{sec:robustness-correlate-performance}

\begin{figure}[!h]
    \centering
    \includegraphics[width=.8\linewidth]{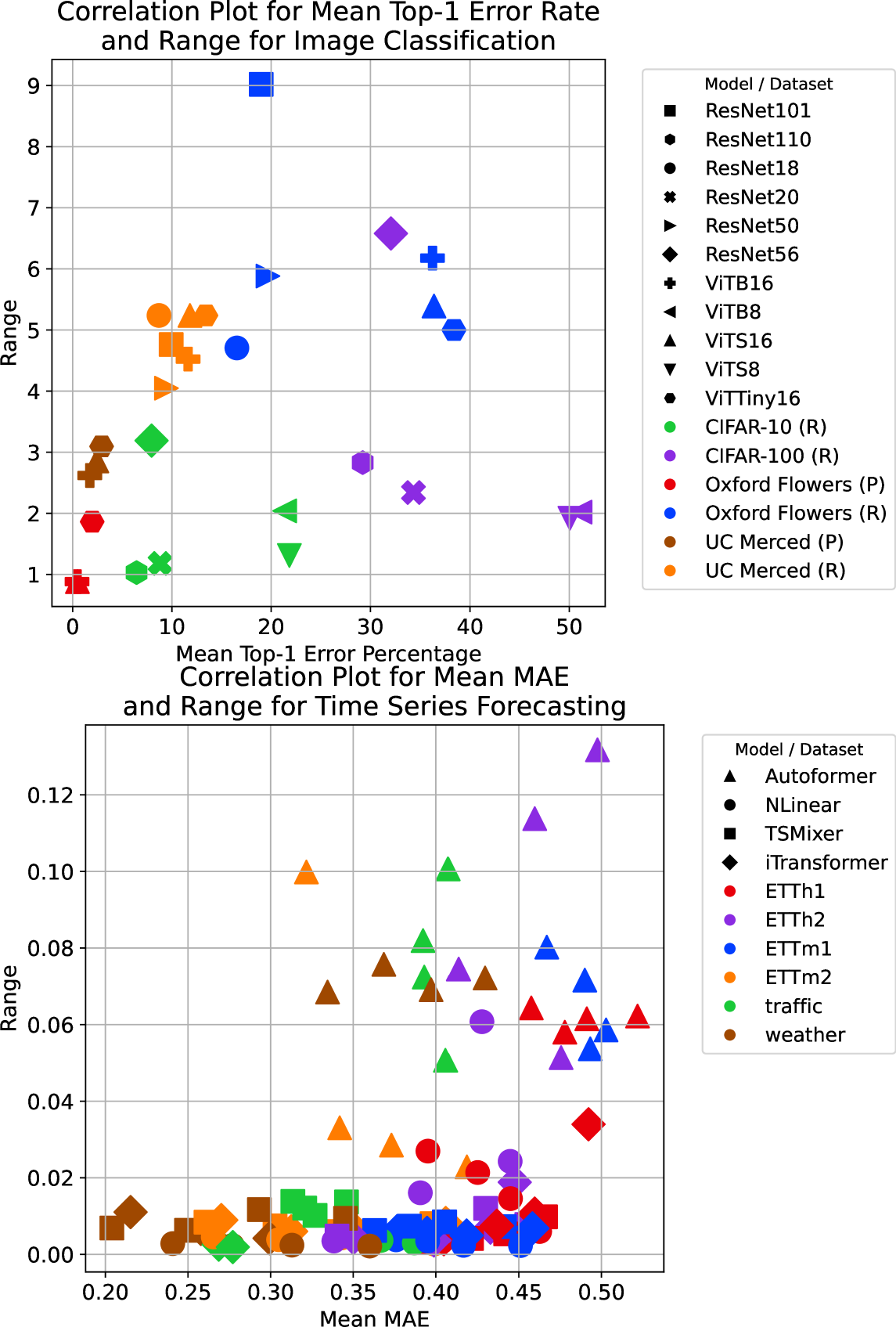}
     \caption{Relationship between mean performance and robustness (range) for image classification (top) and time series forecasting (bottom). Each point represents a model-dataset pair computed from 100 seed replicates. Performance measured as top-1 error for image classification and MAE for time series forecasting. Marker shapes indicate model architecture; colors indicate dataset.}
    \label{figure:correlation_results}
\end{figure}


\begin{table}[t!]
    \centering
    \begin{tabular}{lcccc}
        \toprule
        \multirow{2}{*}{\textbf{Task}} & \textbf{Pearson} & \textbf{p-value} & \textbf{Spearman} & \textbf{p-value} \\
        & \textbf{r} & & \textbf{$\rho$} & \\
        \midrule
        Image Classification    & 0.2053 & 0.2946 & \textbf{0.3750} & \textbf{0.0493} \\
        Time Series Forecasting & \textbf{0.4158} & $\boldsymbol{3.08 \times 10^{-5}}$ & \textbf{0.4211} & $\boldsymbol{2.38 \times 10^{-5}}$ \\
        \bottomrule
    \end{tabular}
    \caption{Pearson and Spearman rank correlation coefficients and p-values for the relationship between mean performance and range (robustness) across tasks. Performance measured as top-1 error (image classification) or MAE (time series forecasting), computed from 100 seed replicates per model-dataset pair. Statistically significant results ($p < 0.05$) in bold.}
    \label{tab:correlation_results}
\end{table}

We examine whether models that achieve better performance also exhibit greater robustness. 
We evaluated this relationship across 122 dataset replicates: 28 from image classification and 94 from time series forecasting. 
Performance was measured using top-1 accuracy for image classification and MAE for time series forecasting. 
Robustness was measured using the range of model performance distribution across 100 seed replicates for each dataset replicate. 
Results are reported in Tables \ref{tab:ic_results} (image classification), \ref{tab:ts_mae_results_1}, \ref{tab:ts_mae_results_2}, and \ref{tab:ts_mae_results_3} (time series forecasting).

To assess the relationship between robustness and performance, we computed both Pearson and Spearman rank correlation coefficients with corresponding p-values between the mean and range of model performance for both tasks.
Pearson's r measures the strength of the linear relationship, while Spearman's $\rho$ measures the monotonic relationship without assuming linearity.
For image classification, we converted top-1 accuracy to top-1 error ($1 - \text{accuracy}$) so that lower values consistently indicate better performance across both tasks.

This analysis quantifies the relationship between performance and robustness across dataset replicates. 
A positive correlation indicates that as performance improves (error decreases), robustness also improves (range decreases). 
Conversely, a negative correlation would indicate that better performance is associated with higher variability. 
We report both Pearson's r and Spearman's $\rho$, each ranging from -1 to +1, where values closer to $\pm1$ indicate stronger relationships. 
We interpret correlation magnitude heuristically using the absolute value of the coefficient: values in $[0.0, 0.3)$ are weak, $[0.3, 0.7)$ are moderate, and $[0.7, 1.0]$ are strong. 
The sign indicates direction, with negative values representing inverse relationships.

The results are presented in \autoref{tab:correlation_results} and visualized in \autoref{figure:correlation_results}. 
For image classification, Pearson's $r = 0.2053$ (p = 0.2946), indicating a weak, non-significant linear relationship. 
Spearman's $\rho$ = 0.375 ($p = 0.0493$); neither survives Holm-Bonferroni correction (Holm-adjusted $p = 0.295$ and $p = 0.099$, respectively). 
Therefore, we cannot conclude that a systematic relationship exists between performance and robustness in image classification after correction.

For time series forecasting, Pearson's r = 0.4158 ($p = 3.08 \times 10^{-5}$; Holm-adjusted $p = 9.5 \times 10^{-5}$) and Spearman's $\rho$ = 0.4211 (p = $2.38 \times 10^{-5}$; Holm-adjusted $p = 9.5 \times 10^{-5}$). 
Both survive Holm-Bonferroni correction, and the close agreement between the two measures confirms that the moderate positive correlation is robust to the linearity assumption. 
Better-performing time series models tend to be more robust, and this relationship is not driven by a small number of influential points.

\autoref{figure:correlation_results} illustrates these patterns visually. 
The scatter plot for image classification shows considerable heterogeneity, consistent with the divergence between Pearson and Spearman results. 
The time series forecasting scatter plot shows a clearer positive trend, where models with lower MAE generally exhibit lower range, and the agreement between correlation measures reinforces this interpretation.

We observed differing patterns across the two tasks. 
In image classification, there was no statistically significant correlation between performance and robustness after Holm-Bonferroni correction.  
In time series forecasting, both measures agreed closely, confirming a moderate and statistically significant positive correlation between performance and robustness.

\subsection{Are Image Classification Models More Robust than Time Series Forecasting Models?}
\label{sec:ic-more-robust-ts}

\begin{table}[t!]
  \centering
  \resizebox{\textwidth}{!}{%
  \begin{tabular}{@{\hspace{6pt}}c|ccc|@{\hspace{10pt}}ccc|@{\hspace{10pt}}cc}
    \toprule
    & \multicolumn{3}{c@{\hspace{10pt}}}{\underline{\textbf{Image Classification}}} & 
    \multicolumn{3}{c@{\hspace{10pt}}}{\underline{\textbf{Time Series Forecasting}}} & 
    \multicolumn{2}{c@{\hspace{10pt}}}{\underline{\textbf{p-values}}} \\
    \textbf{Outlier Type} & \footnotesize\textbf{Outliers} & \footnotesize\textbf{Dataset} & \footnotesize\textbf{Mean} 
    & \footnotesize\textbf{Outliers} & \footnotesize\textbf{Dataset} & \footnotesize\textbf{Mean} & \footnotesize\textbf{Image} & \footnotesize\textbf{Time Series} \\
    &  & \footnotesize\textbf{Replicates} & \footnotesize\textbf{Outliers} &  &  \footnotesize\textbf{Replicates} & \footnotesize\textbf{Outliers} & \footnotesize\textbf{Classification}  & \footnotesize\textbf{Forecasting} \\
    \midrule
    Underperforming & 22 &  22 & 1.0 & 266 &  94 & 2.83 & $\boldsymbol{1.56 \times 10^{-5}}$ & 1.0 \\
    \midrule
    Overperforming & 19 &  22 & 0.86 & 34 &  94 & 0.36 & 0.98 & \textbf{0.02} \\
    \midrule
    Total & 41 &  22 & 1.86 & 300 &  94 & 3.19 & $\boldsymbol{3.26 \times 10^{-3}}$ & 1.0 \\
    \bottomrule
  \end{tabular}
  } 
    \caption{Comparison of outlier counts between image classification and time series forecasting models. Each row shows total outliers, number of dataset replicates, mean outliers per replicate, and p-values for underperforming, overperforming, and total outliers. Welch's t-test used for unequal variances; Student's t-test used for equal variances. Statistically significant results ($p < 0.05$) in bold.}
  \label{tab:outlier-analysis}
\end{table}

We examine whether image classification models are more robust than time series forecasting models using outlier counts as defined in Section \ref{sec:robustness-metrics}. 
Only replicates trained from random weight initialization were included; the six image classification replicates using pretrained weights were excluded to ensure fair comparison. 
Outlier results are reported in Tables \ref{tab:ic_results} (image classification), \ref{tab:ts_mae_results_1}, \ref{tab:ts_mae_results_2}, and \ref{tab:ts_mae_results_3} (time series forecasting).

A key challenge in comparing robustness between tasks is that they use fundamentally different performance metrics. 
Image classification uses top-1 accuracy (bounded, higher is better), while time series forecasting uses MAE (unbounded, lower is better). 
These differences affect tail risk interpretation: underperforming replicates fall in the lower tail for image classification but the upper tail for time series forecasting.

To overcome these differences, we focused on outlier count as a task-agnostic measure of robustness. 
Outlier counts are independent of metric scale because they are based on relative position within a distribution rather than absolute values. 
This allows cross-task comparison without normalization. 
Models producing fewer outliers are considered more robust.

Outliers were grouped into three categories. 
Underperforming outliers were defined as lower fence outliers for image classification and upper fence outliers for time series forecasting. 
Overperforming outliers were defined in reverse. 
Total outliers combined both categories.

For each outlier type, we calculated the mean and variance across all dataset replicates within each task. 
We used Welch's t-test for comparisons with unequal variances (underperforming and total outliers) and independent t-test for equal variances (overperforming outliers).

The results are summarized in \autoref{tab:outlier-analysis}. 
For underperforming outliers, image classification models had substantially fewer on average: 1.0 per replicate compared to 2.83 for time series forecasting. 
This difference was statistically significant and survives Holm-Bonferroni correction ($p = 1.56 \times 10^{-5}$; Holm-adjusted $p = 4.7 \times 10^{-5}$), indicating that image classification models are more robust in avoiding extreme underperformance.

For overperforming outliers, the pattern reversed. 
Image classification replicates averaged 0.86 overperforming outliers compared to 0.36 for time series forecasting. 
This difference was statistically significant and survives correction ($p = 0.02$; Holm-adjusted $p = 0.02$), suggesting that time series forecasting models are less likely to produce exceptional results.

When considering total outliers, image classification models exhibited greater robustness. 
The mean was 1.86 for image classification and 3.19 for time series forecasting. 
This difference was statistically significant and survives correction ($p = 3.26 \times 10^{-3}$; Holm-adjusted $p = 6.52 \times 10^{-3}$), indicating that overall performance is more stable across seed replicates in image classification.

Image classification models are more robust than time series forecasting models. 
Time series forecasting models exhibited nearly three times more underperforming outliers and approximately two-thirds more total outliers on average. 
The overall pattern suggests that performance in image classification is more stable across non-deterministic runs.

\section{Discussion}
\label{sec:discussion}

\subsection{Findings}

Our empirical investigation reveals several key patterns in deep learning robustness to non-determinism:

\begin{description}

    \item[\textbf{Normality (RQ1):}] Model performance distributions are frequently non-Gaussian, particularly in time series forecasting. 
    This pattern is less pronounced but still present in image classification.
    
    \item[\textbf{Model Size (RQ2):}] Model size, measured by parameter count, does not systematically affect robustness. 
    While specific trends emerge within tasks, larger image classification models exhibit fewer outliers, smaller time series models show lower spread, these patterns are inconsistent across robustness metrics, architectures, and tasks.
    
    \item[\textbf{Training Epochs (RQ3):}] Training duration strongly correlates with robustness, but improvement is not continuous. 
    Image classification models generally become more robust with extended training, though the relationship is nonlinear. 
    Time series models often achieve peak robustness early, with degradation upon additional training. 
    Notably, early stopping effectively balances high mean performance with strong robustness, particularly for time series models.
    
    \item[\textbf{Performance-Robustness Relationship (RQ4):}] The correlation between performance and robustness is task-dependent. 
    Time series forecasting exhibits a moderate, statistically significant positive correlation, better-performing models tend to be more robust. 
    No such relationship exists in image classification, where robustness varies independently of mean accuracy.
    
    \item[\textbf{Cross-Task Comparison (RQ5):}] Image classification models demonstrate substantially greater robustness than time series forecasting models. 
    Time series models produce nearly three times more underperforming outliers, indicating higher sensitivity to non-deterministic factors.
    
\end{description}

Two findings warrant particular emphasis. 
First, the prevalence of non-Gaussian distributions (RQ1 \ref{sec:are-mpd-gaussian}) directly undermines the reliability of single-point metrics. 
Non-Gaussian distributions, often skewed or heavy-tailed, can exhibit high mean performance alongside substantial variance or long tails of underperforming outliers. 
This discrepancy between mean performance and robustness (RQ4 \ref{sec:robustness-correlate-performance}) demonstrates that reporting mean accuracy or error alone provides an incomplete and potentially misleading characterization of model quality.

Second, architectural factors show limited systematic influence on robustness. 
Model size does not consistently predict robustness characteristics (RQ2 \ref{sec:does-size-affect-robustness}), and training duration, while correlated with robustness, does not guarantee continuous improvement (RQ3 \ref{sec:increasing-epochs-affect-robustness}). 
These findings suggest that robustness depends on a complex interplay of factors beyond simple architectural choices.

Cross-task analysis reveals that image classification models substantially outperform time series models in robustness (RQ5 \ref{sec:ic-more-robust-ts}). 
This disparity likely reflects differences in research maturity and regularization development between domains, as explored in the following subsection.

\subsection{Task Differences and Regularization}

\begin{figure}[!h]
    \centering
    \includegraphics[width=1\linewidth]{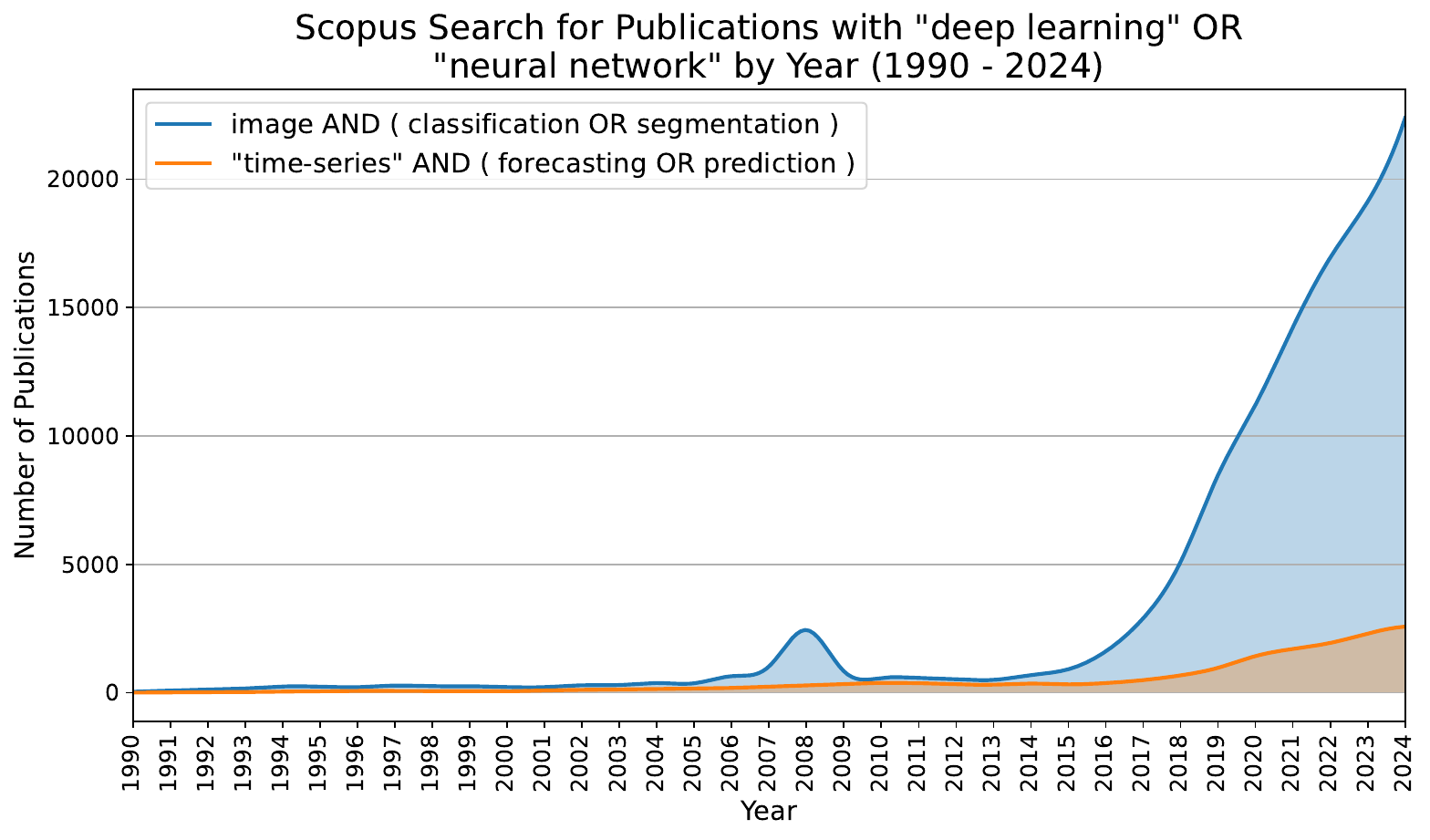}
    \caption{Annual publication counts (1990-2024) for deep learning and neural network papers in image classification/segmentation (blue) versus time series forecasting/prediction (orange) from Scopus. Results show substantially higher research volume in image classification.}
    \label{figure:scopus_image_vs_time}
\end{figure}

The robustness gap between image classification and time series forecasting likely stems from substantial differences in research attention and regularization maturity. 
To quantify this disparity, we analyzed publication volumes using Scopus searches from 1990–2024. Papers containing "deep learning" OR "neural network" alongside "image AND (classification OR segmentation)" numbered 114,319, while those with "time-series AND (forecasting OR prediction)" totaled only 16,673—a nearly sevenfold difference (\autoref{figure:scopus_image_vs_time}).

\begin{figure}[!h]
    \centering
    \includegraphics[width=1\linewidth]{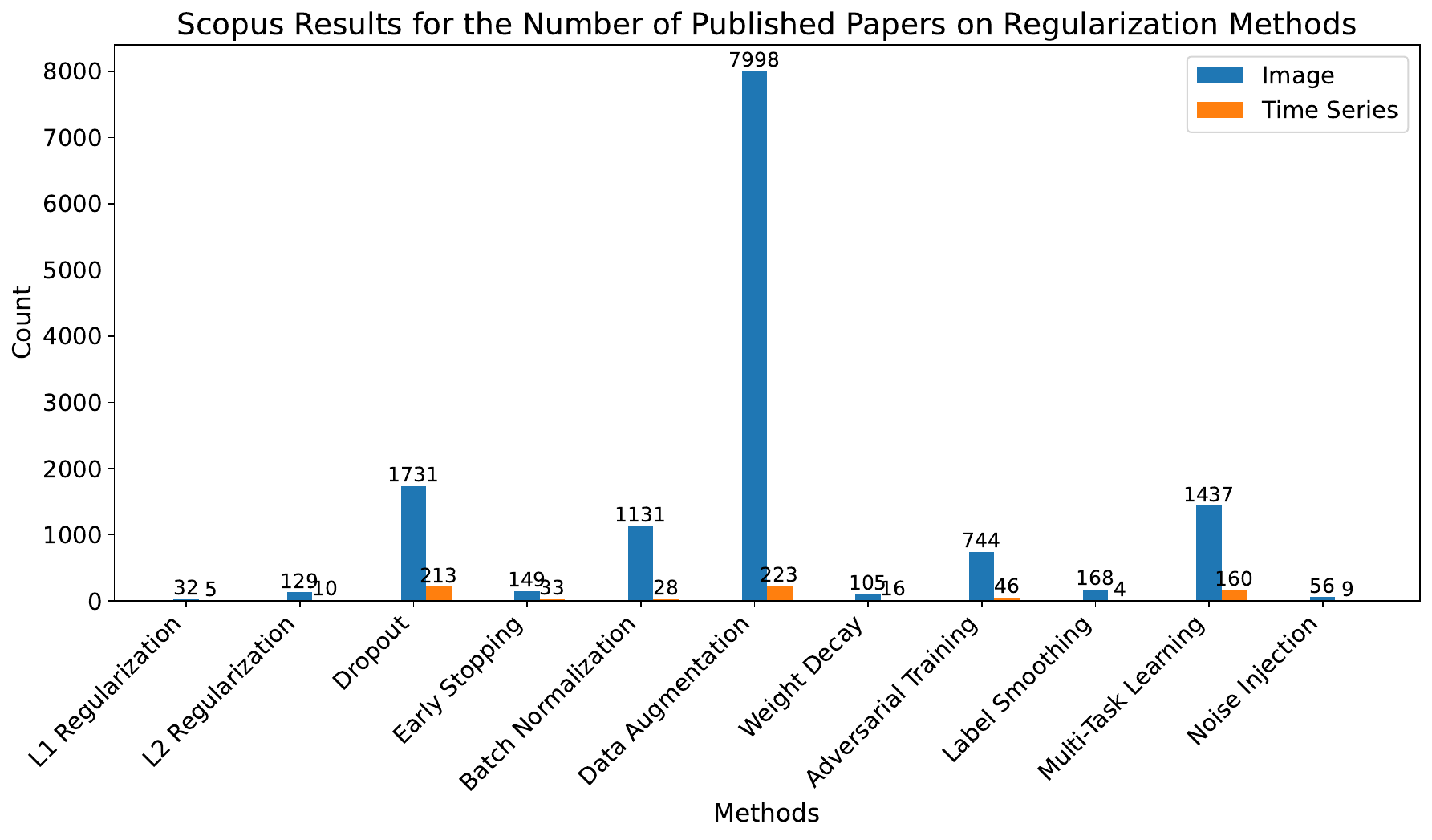}
    \caption{Publication counts for deep learning papers on regularization methods in image classification/segmentation (blue) versus time series forecasting/prediction (orange) from Scopus. All regularization techniques show higher research volume for image classification, with data augmentation being the most studied method.}
    \label{figure:scopus_regularization}
\end{figure}

This imbalance extends to regularization research. 
Following Dietterich's\cite{dietterich_2017} identification of regularization as crucial for robustness, we examined ten widely recognized regularization methods identified across multiple survey papers\cite{kukavcka_2017, nusrat_2018, badola_2020, moradi_2020}. 
Searching for these methods alongside deep learning terms yielded 12,704 papers for image classification versus only 710 for time series forecasting, a difference exceeding 17-fold (\autoref{figure:scopus_regularization}).
Details of the Scopus search method are in \autoref{sec:regularization-method-search}.

\begin{table}[t!]
\centering
\footnotesize
\setlength{\tabcolsep}{3pt} 
\renewcommand{\arraystretch}{1.2} 
\begin{tabularx}{\linewidth}{l X p{1.4cm} p{1.2cm} X X X X X X X}
\toprule
\textbf{Model} & \textbf{L1 Reg} & \textbf{L2 Reg/WD} & \textbf{Dropout} & \textbf{Early Stop} & \textbf{Batch Norm} & \textbf{Data Aug} & \textbf{Adv Train} & \textbf{Label Smooth} & \textbf{MTL} & \textbf{Noise Inj} \\
\midrule
ResNet & - & SGD & - & - & \checkmark & \checkmark & - & - & - & - \\
ViT & - & SGD & \checkmark & - & - & \checkmark & - & - & - & - \\
AutoFormer & - & - & \checkmark & \checkmark & - & - & - & - & - & - \\
iTransformer & - & - & \checkmark & \checkmark & - & - & - & - & - & - \\
Nlinear & - & - & - & \checkmark & - & - & - & - & - & - \\
TSMixer & - & - & \checkmark & \checkmark & \checkmark & - & - & - & - & - \\
\bottomrule
\end{tabularx}
\caption{Regularization techniques used in image classification and time series forecasting models. Column abbreviations: L1 Reg (L1 regularization), L2 Reg/WD (L2 regularization/Weight Decay), Early Stop (Early Stopping), Batch Norm (Batch Normalization), Data Aug (Data Augmentation), Adv Train (Adversarial Training), Label Smooth (Label Smoothing), MTL (Multi-task Learning), Noise Inj (Noise Injection). Checkmarks indicate technique is applied. SGD: models trained with the Stochastic Gradient Descent optimizer use weight decay, models trained with the Adam optimizer do not.}
\label{table:regularization_used}
\end{table}

This research disparity manifests in the models examined in our study. 
Image classification architectures employ specialized regularization techniques including batch normalization and data augmentation—methods specifically developed for imaging tasks. 
Data augmentation, the most extensively researched regularization method, artificially expands training diversity through transformations that preserve spatial relationships while introducing beneficial variation. 
Even basic augmentation methods (random cropping and flipping) used in our experiments provide measurable robustness benefits. 
In contrast, time series models rely predominantly on general-purpose techniques like early stopping, dropout, and weight decay, lacking task-specific adaptations.

The consequences of this gap appear clearly in our results. 
NLinear, which relies solely on early stopping, exhibited the highest frequency of underperforming outliers despite competitive mean performance and spread. 
While early stopping effectively controls overfitting and maintains low spread, it proves insufficient to prevent extreme underperformance. 
This pattern suggests that general-purpose regularization, though valuable, cannot fully address the robustness challenges specific to time series forecasting.

\subsection{Implications for Benchmarking and Reproducibility}

These findings challenge current benchmarking and reproducibility practices. 
Traditional benchmarking emphasizes point estimates—single values such as top-1 accuracy or MAE, often computed from a single run or the mean of fewer than ten runs. 
Our results demonstrate that such metrics obscure the variability induced by non-deterministic factors, yielding potentially misleading conclusions about model superiority.

When performance distributions exhibit wide spread, skewness, or heavy tails, model rankings become unstable, particularly when differences between top performers are small. 
A model appearing superior with one random seed may perform significantly worse with another, a phenomenon we observe across both tasks and previously documented by others\cite{reimers_2017,gundersen_2023}. 
\citet{turan_2026} observe similar ranking instability across different training set sizes and when using different statistical measures (mean versus quantiles) to summarize performance distributions.
This instability undermines the reliability of leaderboards and comparative studies, requiring benchmark-driven claims about state-of-the-art performance to be reconsidered through the lens of distributional robustness rather than mean performance alone.

These issues extend to reproducibility. 
If published results reflect only one or a few replicates from a heavily skewed, long-tailed distribution, independent reproduction efforts may yield inconsistent outcomes not due to methodological error but inherent performance variance. 
Reproducibility efforts should therefore shift from reproducing point estimates to reproducing performance distributions under well-defined protocols involving multiple runs. 
This approach provides a more complete and reliable characterization of model behavior.

\subsection{Trustworthy AI}

Trustworthy AI, as defined by the EU Guidelines\cite{eu_2019}, is built on three pillars: lawfulness, ethical behavior, and robustness. 
Robustness, addressing both technical and social dimensions, is essential to minimize the risk of unintentional harm. 
The notion of Trustworthy AI extends beyond the technical system to encompass the trustworthiness of all processes and actors involved in the AI system's life cycle\cite{kaur2022trustworthy, chatila2021trustworthy}.

Within this framework, robustness is the ability of an AI system to maintain reliable and intended behavior in its operational context, including under conditions of uncertainty and variability\cite{kaur2022trustworthy, chatila2021trustworthy}. 
Technical robustness focuses on ensuring stable performance and minimizing unintended consequences due to system or environmental changes\cite{li2023trustworthy}. 
In this work, we examined robustness to non-determinism by systematically evaluating how deep learning model performance varies when exposed to implementation-level randomness.

Our analysis shows that non-determinism often leads to performance distributions that are non-Gaussian, with substantial spread, skewness, and outliers—features that undermine the reliability and safety of AI systems, particularly in high-stakes applications. 
Furthermore, robustness does not necessarily correlate with mean performance; models with higher mean accuracy may still exhibit substantial variability or tail risk under non-deterministic conditions. 
This highlights the need to evaluate models not only by mean performance, but also by robustness to non-determinism, consistent with priorities identified in recent trustworthy AI literature\cite{li2023trustworthy,kaur2022trustworthy,liang2022advances,chatila2021trustworthy}.

\subsection{ Limitations of the Study and Future Work}

Our investigation into robustness focused specifically on non-determinism arising from stochastic training elements. 
While this significantly influences performance distributions, robustness encompasses broader challenges including adversarial attacks, input perturbations, and distributional shifts\cite{li2023trustworthy,kaur2022trustworthy,liang2022advances,chatila2021trustworthy}. 
Future work should extend this empirical analysis to these additional dimensions and assess whether similar distributional patterns emerge.

The selection of architectures and datasets was necessarily constrained by computational resources.
We chose representative models and datasets to enable extensive analysis, but observed patterns may not universally generalize to all architectures, datasets, or application domains. 
Future research should expand coverage to assess generalizability more thoroughly.

Computational budget constraints significantly influenced experimental design. 
Training each model one hundred times to characterize performance distributions is computationally intensive, necessitating careful balancing of analytical depth against resource allocation. 
This limited the number of architectures, datasets, and hyperparameters explored. 
Increased computational resources would enable more extensive investigations into deep learning robustness.

Finally, while this study examines non-determinism from stochastic elements, other sources of irreproducibility\cite{gundersen_2022sources}, including hyperparameter optimization and dataset preprocessing variations, also affect performance distributions but were not our primary focus. 
We identify characterizing robustness across the hyperparameter space as an important direction for future work.
Future work should investigate the combined or individual effects of these non-stochastic factors on model performance variability.

\section{Conclusion}
\label{sec:conclusion}

This empirical study demonstrates that deep learning model performance is best understood as a distribution rather than a fixed value, with variability arising from non-deterministic factors during training. 
Through systematic evaluation using 100 seed replicates per model across 186 experiments, we establish robust metrics of spread, symmetry, and tail risk that characterize performance distributions and enable reliable robustness assessment. 
Spread metrics (range and standard deviation) quantify overall variability, symmetry metrics (skewness) identify whether distributions are skewed toward underperformance or overperformance, and tail risk metrics (outlier counts via Tukey's fences) capture extreme failures that pose the greatest risk in high-stakes applications. 
Our work provides the first cross-task comparison of robustness using metrics independent of task-specific performance measures, enabling direct assessment of distributional characteristics across image classification and time series forecasting.

Our findings reveal critical patterns in how robustness varies across architectures and tasks. 
Consistent with prior work, we confirm that performance distributions are frequently non-Gaussian, with this pattern significantly more pronounced in time series forecasting than image classification. 
Model size does not systematically improve robustness: while larger image classification models exhibited fewer outliers, smaller time series models consistently demonstrated lower spread. 
Training duration does not consistently improve robustness—while image classification models generally became more robust with extended training, robustness did not improve continuously, and time series models often achieved peak robustness early, with degradation upon further training. 
Early stopping effectively balances high mean performance with strong robustness across both tasks. Most critically, mean performance does not predict robustness: time series forecasting showed only moderate correlation, while image classification showed no significant relationship. 
This disconnect confirms that reporting mean accuracy or error alone provides incomplete characterization, masking substantial variability and tail risk in performance distributions.

Cross-task analysis reveals substantial differences in distributional robustness. 
Time series forecasting models produced nearly three times more underperforming outliers than image classification models, indicating greater susceptibility to catastrophic failures from non-deterministic factors—a risk obscured by mean performance metrics. 
The robustness gap likely reflects differences in research maturity: image classification benefits from specialized regularization techniques like batch normalization and data augmentation, supported by over seventeen times more publications on regularization methods, while time series models rely predominantly on general-purpose techniques. 
This disparity translates into measurably higher tail risk and more asymmetric performance distributions in time series forecasting.

These findings challenge current practices in benchmarking and reproducibility. 
Single-run or small-sample evaluations obscure variability from non-determinism, potentially yielding unstable model rankings. 
Reproducibility efforts must shift from reproducing point estimates to characterizing performance distributions through standardized protocols with sufficient seed replicates. For Trustworthy AI, robustness to non-determinism is essential for reliability. 
In high-stakes domains—medical diagnostics, autonomous systems, financial modeling—tail risk poses unacceptable consequences: models performing well on average may exhibit catastrophic failures that are only revealed through distributional analysis. 
Standard evaluation protocols must systematically assess spread, symmetry, and tail risk, not merely mean performance. 
Only by characterizing complete performance distributions can we reliably assess model behavior where consistent performance is critical.

\backmatter








\section*{Declarations}

\subsection{Funding}

The authors received no financial support for the research, authorship, and/or publication of this article.

\subsection{Ethics Approval}

This article does not contain any studies with human or animal participants.

\subsection{Informed Consent}

There are no human participants in this article and informed consent is not required.

\subsection{Author Contributions}

K.L.C. conceived and designed the experiments, performed the experiments, analyzed
the data, contributed materials/analysis tools, and wrote the paper. 
O.E.G. conceived and designed the experiments, analyzed the data, contributed
materials/analysis tools, and wrote the paper.

\subsection{Data Availability}

The training data and results can be found online at \href{https://github.com/kevincoakley/measuring-deep-learning-performance}{https://github.com/kevincoakley/measuring-deep-learning-performance}.

\subsection{Code Availability}

The code for training and analysis can be found online at \href{https://github.com/kevincoakley/measuring-deep-learning-performance}{https://github.com/kevincoakley/measuring-deep-learning-performance}.

\subsection{Conflict of Interest}

The authors declare no conflict of interest.

\subsection{Competing Interest}

The authors declare no competing interests.

\subsection{Clinical Trial Number}

Not applicable.


\begin{appendices}

\section{Experiment Details}
\label{sec:apx-experiment-details}

\subsection{Image Classification}\label{secA1}

\subsubsection{Datasets, Data Splits, Preprocessing, and Augmentation}

\textbf{Datasets:} For the image classification experiments we use four datasets, two datasets with 32x32x3 pixel images and two datasets with 256x256x3 pixel images.
For the 32x32x3 pixel images, the CIFAR-10 and CIFAR-100\cite{krizhevsky_09} datasets were used. 
CIFAR-10 includes 60,000 images across 10 classes, while CIFAR-100 also contains 60,000 images of the same resolution, but categorized into 100 classes for a more detailed classification.
For the 256x256x3 pixel images the Oxford Flowers\cite{nilsback_08} and UC Merced\cite{yang_2010} datasets were used. 
Oxford Flowers comprises 8,189 images of varying dimensions across 102 classes, and UC Merced features 2,100 images divided into 21 classes.

\textbf{Data Splits:} In our experiments, the CIFAR-10 and CIFAR-100 datasets were divided into train, validation, and test sets containing 45,000, 5,000, and 10,000 images respectively, following the instructions by He et al.\cite{he_2016}. 
The Oxford Flowers dataset features a test set of 6,149 images and equal-sized training and validation sets of 1,020 images each. 
To improve accuracy, the test set was repurposed for training, and the training set for testing. 
For the UC Merced dataset, which does not have predefined splits, we applied a 70\%/10\%/20\% split for training, validation, and testing.

Image allocation to each split in the CIFAR-10, CIFAR-100, and UC Merced datasets was achieved through a one-time random selection process. 
We ensured a uniform distribution of images across labels within these splits. 
Consistency was maintained by allocating the same images to the same splits in every training run.

\textbf{Preprocessing:} Image preprocessing involved normalizing the color channels based on the mean and standard deviation calculated across the entire dataset. 
For Oxford Flowers and UC Merced, images were further processed by resizing to 224x224x3, employing "nearest" interpolation.

\textbf{Augmentation:} Data augmentation techniques were consistently applied across all training datasets in our experiments, following the methods outlined by Lee et al.\cite{lee_2015} and recommended by He et al., to improve model robustness and generalization.

\subsubsection{Models}

\begin{table*}[t]
    \centering
    \begin{tabular}{c|c|c}
    \toprule
      Model & Variation & Parameters \\
    \hline
    \multirow{6}{*}{ResNet}
      & 20 & 0.27M \\
      & 56 & 0.85M \\
      & 110 & 1.73M \\
      & 18 & 11.18M \\ 
      & 50 & 23.52M \\
      & 101 & 42.52M \\
    \midrule
    \multirow{3}{*}{ViT}
     & Tiny & 5.53M \\
     & Small & 21.67M \\
     & Base & 85.81M \\
    \bottomrule
    \end{tabular}
    \caption{The number of parameters for each of the Image Classification models.}
    \label{table:image_classification_parameters}
\end{table*}

The ResNet models were developed using the PyTorch framework, following the protocols specified by He et al. 
The 20, 56, and 110 layer versions were used for the 32x32x3 pixel images, while the 18, 50, and 101 layer versions were used for 256x256x3 pixel images.

We used the ViT models from Hugging Face's Timm library\footnote{https://github.com/huggingface/pytorch-image-models}. 
We used the ViT-Small\cite{steiner_2021} (ViT-S), and ViT-Base\cite{dosovitskiy_2020} (ViT-B) variations with patch size 8 for the 32x32x3 pixel images. 
We used the ViT-Tiny\cite{steiner_2021} (ViT-T), ViT-Small (ViT-S), and ViT-Base (ViT-B) with variations with patch size 16 for the 256x256x3 pixel images.
Diverging from the approach in the original paper by Dosovitskiy et al., these experiments performed model training with random weight initialization rather than using pre-trained weights from the ImageNet and JFT datasets. 
This difference led to top-1 accuracy metrics that are notably lower compared to those achieved by the ResNet models and those reported by Dosovitskiy et al., who used pre-trained weights. 
Dosovitskiy et al. noted the challenges Vision Transformers face in generalizing effectively with limited data, recommending training on substantially larger datasets (ranging from 14M to 300M images) to achieve optimal performance.

ViT models were also trained using pretrained weights from the Timm library. 
These pretrained weights were not used with the 32x32x3 pixel images due to the requirement of image dimensions of 224x224x3.

See table \ref{table:image_classification_parameters} for the number of parameters in each model.

\subsubsection{Hyperparameters}

\begin{table*}[t]
    \centering
    \begin{tabular}{c|c|c|c|c}
    \toprule
        Model & Dataset  & Learning Rate & Batch Size & Optimizer \\
        \hline
        \multirow{4}{*}{ResNet} 
        & CIFAR-10 & 0.1 & 128 & SGD \\
        & CIFAR-100 & 0.1 & 128 & SGD \\
        & Oxford Flowers & 0.001 & 64 & Adam \\
        & UC Merced & 0.001 & 32 & Adam \\
        \midrule
        \multirow{4}{*}{ViT (Random Weights)}
         & CIFAR-10 & 0.0001 & 128 & Adam \\
         & CIFAR-100 & 0.0001 & 128 & Adam \\
         & Oxford Flowers & 0.0001 & 128 & Adam \\
         & UC Merced & 0.001 & 128 & Adam \\
        \midrule
        \multirow{4}{*}{ViT (Pretrained Weights)}
         & CIFAR-10 & 0.001 & 128 & SGD \\
         & CIFAR-100 & 0.001 & 128 & SGD \\
         & Oxford Flowers & 0.001 & 128 & SGD \\
         & UC Merced & 0.001 & 128 & SGD \\
        \bottomrule
    \end{tabular}
    \caption{Hyperparameters selected for training based on a grid search.}
    \label{table:image_classification_hyperparameters}
\end{table*}

For experiments involving ResNet models with the CIFAR-10 and CIFAR-100 datasets, we adhered to the hyperparameters specified by He et al. in section 4.2 of their publication.

For other model and dataset configurations, hyperparameters were optimized through a systematic grid search, covering variations in learning rate, batch size, and optimizer choice. 
Learning rates tested were 0.1, 0.01, 0.001, and 0.0001; batch sizes considered were 32, 64, 128, and 256; and the optimizers evaluated included Stochastic Gradient Descent (SGD) and Adam.

Additionally, all models incorporated learning rate decay, with a decay factor of 0.1 applied after 50\% and 75\% of the total epochs were completed. 
The ResNet 110 models implemented a five-epoch learning rate warm-up at 0.1, following He et al.'s recommendation, due to the model's initial difficulty in converging.

Models were trained for 200 epochs using random weights and 50 epochs with pretrained weights.

See table \ref{table:image_classification_hyperparameters} for the hyperparameters used.

\subsection{Time Series Forecasting}
\label{sec:appendix-time-series}

\subsubsection{Datasets}

In our time series experiments, we utilized six datasets:

The four ETT (Electricity Transformer Temperature) datasets\cite{zhou_2021} are aimed at evaluating forecasting models for monitoring transformer temperatures. 
ETTh1 and ETTh2 feature hourly data with 7 variates across 17,420 timesteps. 
ETTm1 and ETTm2 provide data every 5 minutes, also with 7 variates, across 69,680 timesteps.

The Traffic dataset, sourced from the California Department of Transportation, records sensor data on freeways in the San Francisco Bay area, with hourly granularity, 862 variates, and 17,544 timesteps. 
The Weather dataset comprises various indicators recorded in Germany in 2020, detailed every 10 minutes, with 21 variates and 52,696 timesteps. 
Both the Traffic and Weather datasets were preprocessed as described by Wu et al\cite{wu_2021}.

Forecasting horizons of 96, 192, 336, and 720 time steps were used for iTransformer, NLinear, and TSMixer models, aligning with published results. 
To maintain consistency across all models, these same horizons were applied to the Autoformer, deviating from the original authors' approach. 
A uniform sequence length of 96 was utilized for all four models, corresponding to the published parameters for Autoformer and iTransformer. 
This standardization ensures comparability of results across the different models.

\subsubsection{Models and Hyperparameters}

\begin{table*}[t]
    \centering
    \begin{tabular}{c|c|c|c|c|c}
    \toprule
      Dataset & Horizon & Autoformer & iTransformer & NLinear & TSMixer \\
    \hline
    \multirow{4}{*}{ETTh1} 
        & 96  & 10.54M & 0.84M & 0.01M & 0.04M \\
        & 192 & 10.54M & 0.86M & 0.02M & 0.05M \\
        & 336 & 10.54M & 0.90M & 0.03M & 0.06M \\
        & 720 & 10.54M & 1.00M & 0.07M & 0.10M \\ 
    \hline
    \multirow{4}{*}{ETTh2} 
        & 96 & 10.54M & 0.22M & 0.01M & 0.04M \\
        & 192 & 10.54M & 0.24M & 0.02M & 0.05M \\
        & 336 & 10.54M & 0.26M & 0.03M & 0.06M \\
        & 720 & 10.54M & 0.30M & 0.07M & 0.10M \\
    \hline
    \multirow{4}{*}{ETTm1} 
        & 96 & 10.54M & 0.22M & 0.01M & 0.04M \\
        & 192 & 10.54M & 0.24M & 0.02M & 0.05M \\
        & 336 & 10.54M & 0.26M & 0.03M & 0.06M \\
        & 720 & 10.54M & 0.30M & 0.07M & 0.10M \\
    \hline
    \multirow{4}{*}{ETTm2} 
        & 96 & 10.54M & 0.22M & 0.01M & 0.04M \\
        & 192 & 10.54M & 0.24M & 0.02M & 0.05M \\
        & 336 & 10.54M & 0.26M & 0.03M & 0.06M \\
        & 720 & 10.54M & 0.30M & 0.07M & 0.10M \\
    \hline
    \multirow{4}{*}{Traffic} 
        & 96 & 14.91M & 6.41M & 0.01M & 6.27M \\
        & 192 & 14.91M & 6.46M & 0.02M & 6.28M \\
        & 336 & 14.91M & 6.53M & 0.03M & 6.30M \\
        & 720 & 14.91M & 6.73M & 0.07M & 6.33M \\
    \hline
    \multirow{4}{*}{Weather} 
        & 96 & 10.61M & 4.83M & 0.01M & 0.12M \\
        & 192 & 10.61M & 4.88M  & 0.02M & 0.13M \\
        & 336 & 10.61M & 4.96M & 0.03M & 0.14M \\
        & 720 & 10.61M & 5.15M & 0.07M & 0.18M \\
    \bottomrule
    \end{tabular}
    \caption{The number of parameters for each of the Time Series models.}
    \label{table:time_series_parameters}
\end{table*}

All models utilized in our study were trained using code provided by the authors, available on GitHub:

\begin{itemize}
    \item Autoformer: https://github.com/thuml/autoformer
    \item iTransformer: https://github.com/thuml/iTransformer
    \item NLinear: https://github.com/cure-lab/LTSF-Linear
    \item TSMixer: https://github.com/google-research/google-research
\end{itemize}

Each GitHub repository contained scripts to run the experiments, including predefined hyperparameters as documented in their respective publications. 
Modifications were made to the original code to ensure deterministic training, setting random seeds, and saving the outcomes.

We considered CNN time series models, like SCINet\cite{liu_2022} and TiDE\cite{das_2023}, however we chose not to use them in our experiments due to the long training times.

See table \ref{table:time_series_parameters} for the number of parameters in each model.

\subsection{Training}
\label{sec:appendix-training}

All Image Classification and Time Series models were trained deterministically one hundred times using the same one hundred randomly selected seeds. 
Deterministic training was enforced through framework-specific mechanisms for PyTorch and TensorFlow. For PyTorch, \texttt{torch.use\_deterministic\_algorithms(True)} was called prior to each run, forcing all applicable CUDA operations - including convolutions, scatter operations, and matrix multiplications - to use deterministic algorithms and raising a \texttt{RuntimeError} if no deterministic implementation is available\footnote{\href{https://docs.pytorch.org/docs/2.2/generated/torch.use_deterministic_algorithms.html}{https://docs.pytorch.org/docs/2.2/generated/torch.use\_deterministic\_algorithms.html}}.
The \texttt{CUBLAS\_WORKSPACE\_CONFIG=:4096:8} environment variable was also set to ensure deterministic cuBLAS behavior, as required for CUDA versions 10.2 and above. cuDNN autotuning was not enabled (\texttt{torch.backends.cudnn.benchmark} defaults to \texttt{False} in PyTorch 2.2), preventing hardware-dependent algorithm selection across runs.
For TensorFlow, \texttt{tf.config.experimental.enable\_op\_determinism()} was called to enforce deterministic op execution\footnote{\href{https://www.tensorflow.org/versions/r2.14/api_docs/python/tf/config/experimental/enable_op_determinism}{https://www.tensorflow.org/versions/r2.14/api\_docs/python/tf/config/experimental/enable\_op\_determinism} }, and \texttt{PROTOCOL\_BUFFERS\_PYTHON\_IMPLEMENTATION=python} was set to ensure consistent protobuf serialization behavior across nodes. 
In both frameworks, the random states of Python, NumPy, and the respective framework were seeded identically at the start of every run. 
Together, these settings ensure that observed variation across the 100 seed replicates reflects seed-based non-determinism only, and not uncontrolled implementation-level sources of variation.

When using the HPC system for training, deterministic results across each node were verified by training the models for one epoch using the same random seed and validating the top-1 accuracy results were the same.

\subsection{Software Environment}

All the experiments were conducted within a software environment derived from Nvidia NGC containers available from https://docs.nvidia.com/deeplearning/frameworks/. 
To accommodate the HPC user security requirements, the Docker containers were converted to Apptainer containers. 
Version 23.12 of the TensorFlow and PyTorch containers was utilized. 
The TensorFlow container includes TensorFlow version 2.14.0, while the PyTorch container includes PyTorch version 2.2.0. 
Both containers are built on an Ubuntu 22.04 base and incorporate Python 3.10, CUDA 12.3.2, and cuDNN 8.9.7.29.

\section{Regularization Method Search}
\label{sec:regularization-method-search}

Regularization is a method identified by Dietterich\cite{dietterich_2017} to increase robustness of machine learning models.
There exists little published research on regularization methods developed specifically for time series forecasting methods.
We reviewed four survey papers \cite{kukavcka_2017, nusrat_2018, badola_2020, moradi_2020} on regularization methods for deep learning and selected ten regularization methods that were listed in two or more of the survey papers.  
We then did a Scopus search for publications with "deep learning" OR "neural network" and the ten regularization methods: 
"L1 regularization"\cite{kukavcka_2017, nusrat_2018, badola_2020, moradi_2020} 
OR "L2 regularization" OR "weight decay" \cite{kukavcka_2017, nusrat_2018, badola_2020, moradi_2020} 
OR "dropout"\cite{kukavcka_2017, nusrat_2018, badola_2020, moradi_2020} 
OR "early stopping"\cite{kukavcka_2017, nusrat_2018, badola_2020, moradi_2020} 
OR "batch normalization"\cite{kukavcka_2017, nusrat_2018, badola_2020, moradi_2020} 
OR "data augmentation"\cite{kukavcka_2017, nusrat_2018, badola_2020, moradi_2020} 
OR "adversarial training"\cite{kukavcka_2017, moradi_2020} 
OR "label smoothing"\cite{kukavcka_2017, moradi_2020} 
OR "multi-task learning" OR "multitask learning"\cite{kukavcka_2017, moradi_2020} 
OR "noise injection"\cite{kukavcka_2017, moradi_2020} 
returned 12704 papers with "image AND (classification OR segmentation)" in the title, abstract, or keywords, while only 710 papers had "time-series AND (forecasting OR prediction)," see Figure \ref{figure:scopus_regularization}.

\section{Additional Tables}

\subsection{Image Classification}

\begin{table}[h!]
\centering
\resizebox{\textwidth}{!}{%
\begin{tabular}{lc|c|c|cc|ccc|cc}
\hline
\multirow{2}{*}{} & & \multirow{1}{*}{} & \multirow{1}{*}{} & \multicolumn{2}{c|}{Spread} & \multicolumn{3}{c|}{Symmetry} & \multicolumn{2}{c}{Tail Risk} \\
Model & Dataset & \# Para & Mean & Range & SD & Skewness & Shapiro W & Shapiro p & LFO & UFO \\
\midrule
\multicolumn{10}{l}{\textit{32x32x3 Pixel Images \& Random Weight Initialization}} \\
\midrule
ResNet20 & 
\multirow{5}{*}{CIFAR-10} 
         & 0.27M & 91.19\% & 1.18\% & 0.22 & -0.4176 & 0.9802 & 0.1382 & 3 & 0 \\
ResNet56 & & 0.85M & 92.05\% & 3.19\% & 0.61 & -0.9379 & 0.9475 & 0.0006 & 5 & 0 \\
ResNet110 & & 1.73M & 93.56\% & 1.03\% & 0.17 & 0.3784 & 0.9783 & 0.0981 & 0 & 2 \\
ViT-S-8 & & 21.67M & 78.16\% & 1.31\% & 0.27 & -0.4026 & 0.9747 & 0.0507 & 0 & 0 \\
ViT-B-8 & & 85.81M & 78.64\% & 2.04\% & 0.32 & 0.2893 & 0.9711 & 0.0269 & 1 & 3 \\
\midrule
ResNet20 & 
\multirow{5}{*}{CIFAR-100} 
         & 0.27M & 65.66\% & 2.34\% & 0.44 & 0.1881 & 0.9882 & 0.5227 & 0 & 2 \\
ResNet56 & & 0.85M & 67.94\% & 6.58\% & 1.41 & -0.8397 & 0.9371 & 0.0001 & 1 & 0 \\
ResNet110 & & 1.73M & 70.77\% & 2.83\% & 0.49 & -0.642 & 0.9679 & 0.0153 & 2 & 0 \\
ViT-S-8 & & 21.67M & 49.94\% & 1.92\% & 0.36 & 0.0326 & 0.9906 & 0.7111 & 0 & 2 \\
ViT-B-8 & & 85.81M & 48.92\% & 2.02\% & 0.45 & -0.179 & 0.9892 & 0.6 & 0 & 0 \\
\midrule
\multicolumn{10}{l}{\textit{256x256x3 Pixel Images \& Random Weight Initialization}} \\
\midrule 
ResNet18 &
\multirow{6}{*}{Oxford Flowers} 
         & 11.18M & 83.45\% & 4.71\% & 0.97 & -0.2553 & 0.989 & 0.5876 & 2 & 0 \\
ResNet50 & & 23.52M & 80.32\% & 5.88\% & 1.13 & -0.1544 & 0.9938 & 0.9314 & 2 & 0 \\
ResNet101 & & 42.52M & 81\% & 9.02\% & 1.78 & -0.6234 & 0.9663 & 0.0117 & 1 & 0 \\
ViT-T-16 & & 5.53M & 61.6\% & 5\% & 1.08 & 0.3233 & 0.9821 & 0.194 & 0 & 2 \\
ViT-S-16 & & 21.67M & 63.59\% & 5.39\% & 1.14 & 0.3973 & 0.9801 & 0.1363 & 0 & 4 \\
ViT-B-16 & & 85.81M & 63.74\% & 6.18\% & 1.16 & -0.2817 & 0.9736 & 0.0422 & 1 & 1 \\
\midrule 
ResNet18 &
\multirow{6}{*}{UC Merced} 
        & 11.18M & 91.3\% & 5.24\% & 0.93 & 0.1171 & 0.9883 & 0.5342 & 1 & 2 \\
ResNet50 & & 23.52M & 90.55\% & 4.05\% & 0.93 & -0.1839 & 0.9726 & 0.0348 & 0 & 0 \\
ResNet101 & & 42.52M & 90.07\% & 4.76\% & 1.03 & 0.2944 & 0.9772 & 0.0809 & 0 & 0 \\
ViT-T-16 & & 5.53M & 86.57\% & 5.24\% & 0.96 & -0.465 & 0.9781 & 0.0937 & 2 & 0 \\
ViT-S-16 & & 21.67M & 88.18\% & 5.24\% & 0.93 & 0.2405 & 0.9864 & 0.3986 & 1 & 1 \\
ViT-B-16 & & 85.81M & 88.37\% & 4.52\% & 1.03 & -0.0483 & 0.9832 & 0.2337 & 0 & 0 \\
\midrule 
\multicolumn{10}{l}{\textit{256x256x3 Pixel Images \& Pretrained Weight Initialization}} \\
\midrule
ViT-T-16 & 
\multirow{3}{*}{Oxford Flowers}
         & 5.53M & 98.03\% & 1.86\% & 0.33 & -0.5009 & 0.9738 & 0.0439 & 1 & 1 \\
ViT-S-16 & & 21.67M & 99.52\% & 0.88\% & 0.18 & -0.4222 & 0.9589 & 0.0034 & 2 & 0 \\
ViT-B-16 & & 85.81M & 99.56\% & 0.88\% & 0.18 & -0.2033 & 0.9622 & 0.0058 & 0 & 0 \\
\midrule
ViT-T-16 & 
\multirow{3}{*}{UC Merced}
         & 5.53M & 97.08\% & 3.1\% & 0.64 & 0.1522 & 0.9816 & 0.1766 & 0 & 0 \\
ViT-S-16 & & 21.67M & 97.56\% & 2.86\% & 0.58 & -0.6398 & 0.9558 & 0.0021 & 3 & 0 \\
ViT-B-16 & & 85.81M & 98.27\% & 2.62\% & 0.45 & 0.4513 & 0.9639 & 0.0077 & 1 & 1 \\
\hline
\end{tabular}
} 
\caption{Performance and robustness metrics for image classification models. Each row shows results for 100 seed replicates. Columns: \# Para (parameter count), Mean (mean top-1 accuracy), Range, SD (standard deviation), Skewness, Shapiro W (Shapiro-Wilk W statistic), Shapiro p (Shapiro-Wilk p-value), LFO (lower fence outliers), UFO (upper fence outliers). Models grouped by image resolution and initialization strategy.}
\label{tab:ic_results}
\end{table}

\begin{table}[!t]
\centering
\resizebox{\textwidth}{!}{%
\footnotesize
\setlength{\tabcolsep}{4pt}
\begin{tabular}{lccccc}
\toprule
\textbf{Robustness Measure} & \textbf{W} & \textbf{p-value} & \textbf{r} & \textbf{More Robust} & \textbf{Effect Size} \\
\midrule
\multicolumn{6}{l}{\textit{Intra-architecture Wilcoxon Comparison}} \\
\midrule
Range                & 155.5 & 0.6204          & \phantom{$-$}0.114  & ---    & Small \\
Standard Deviation   & 135.0 & 0.3153          & \phantom{$-$}0.231  & ---    & Small \\
Skewness             & 163.0 & 0.7644          & \phantom{$-$}0.071  & ---    & Negligible \\
Lower Fence Outliers & 107.0 & 0.5240          & $-$0.154            & ---    & Small \\
Upper Fence Outliers & 63.0  & 0.5301          & $-$0.177            & ---    & Small \\
All Outliers         & 74.0  & 0.2479          & $-$0.295            & ---    & Small \\
\midrule
\multicolumn{6}{l}{\textit{Cross-architecture Wilcoxon Comparison}} \\
\midrule
Range                & 732.5 & 0.7565          & $-$0.049            & ---    & Negligible \\
Standard Deviation   & 686.0 & 0.3631          & \phantom{$-$}0.140  & ---    & Small \\
Skewness             & 564.0 & 0.0568          & $-$0.293            & ---    & Small \\
Lower Fence Outliers & 333.5 & \textbf{0.0337} & $-$0.356            & Larger & Medium \\
Upper Fence Outliers & 328.0 & 0.5355          & $-$0.115            & ---    & Small \\
All Outliers         & 235.5 & \textbf{0.0062} & $-$0.478            & Larger & Medium \\
\bottomrule
\end{tabular}
} 
\caption{Wilcoxon signed-rank test results for robustness comparisons between image classification models with different parameter counts. W Statistic and p-value report the Wilcoxon signed-rank test statistic and corresponding p-value for paired comparisons on raw metric values. r denotes the rank-biserial correlation coefficient. More Robust indicates which model size exhibited lower metric values on the robustness measure; --- indicates no statistically significant difference. Effect Size reports the rank-biserial correlation magnitude using the thresholds: negligible ($|r| < 0.1$), small ($0.1 \leq |r| < 0.3$), medium ($0.3 \leq |r| < 0.5$), and large ($|r| \geq 0.5$). Statistically significant results ($p < 0.05$) in bold.}
\label{tab:ic_wilcoxon}
\end{table}

\subsection{Time Series Forecasting}

\clearpage
\begin{table}[h!]
\centering
\resizebox{\textwidth}{!}{%
\begin{tabular}{lcc|c|c|cc|ccc|cc}
\hline
\multirow{3}{*}{} & & & \multirow{1}{*}{} & \multirow{1}{*}{} & \multicolumn{2}{c|}{Spread} & \multicolumn{3}{c|}{Symmetry} & \multicolumn{2}{c}{Tail Risk} \\
Model & Dataset & Horizon & \# Para & Mean & Range & SD & Skewness & Shapiro W & Shapiro p & LFO & UFO \\
\midrule
Autoformer & 
\multirow{4}{*}{ETTh1} &
\multirow{4}{*}{96} 
         & 10.54M & 0.4576 & 0.0644 & 0.0141 & 0.7442 & 0.9494 & 0.0008 & 0 & 5 \\
iTransformer & & & 0.84M & 0.4049 & 0.004 & 0.0009 & 0.2521 & 0.9768 & 0.0753 & 0 & 0 \\
NLinear & & & 0.01M & 0.3952 & 0.0269 & 0.004 & 3.447  & 0.6492 & $4.09 \times 10^{-4}$ & 0 & 5 \\
TSMixer & & & 0.04M & 0.3921 & 0.0033 & 0.0008 & 0.1937 & 0.9775 & 0.085 & 0 & 0 \\
\midrule
Autoformer &
\multirow{4}{*}{ETTh1} &
\multirow{4}{*}{192} 
         & 10.54M & 0.4778 & 0.058 & 0.0116 & 0.1648 & 0.9884 & 0.5362 & 1 & 0 \\
iTransformer & & & 0.86M & 0.4367 & 0.0075 & 0.0015 & 0.4386 & 0.9741 & 0.0456 & 0 & 0 \\
NLinear & & & 0.02M & 0.4251 & 0.0214 & 0.0038 & 2.1074 & 0.7715 & $3.60 \times 10^{-11}$ & 0 & 5 \\
TSMixer & & & 0.05M & 0.4214 & 0.0041 & 0.0008 & 0.5489 & 0.9748 & 0.0524 & 0 & 2 \\
\midrule
Autoformer &
\multirow{4}{*}{ETTh1} &
\multirow{4}{*}{336} 
         & 10.54M & 0.4911 & 0.0616 & 0.012 & 1.1397 & 0.9195 & $1.33 \times 10^{-5}$ & 0 & 5 \\
iTransformer & & & 0.9M & 0.4596 & 0.0106 & 0.0023 & 0.066 & 0.9918 & 0.8049 & 0 & 0 \\
NLinear & & & 0.02M & 0.445 & 0.0146 & 0.0026 & 2.4869 & 0.7117 & $1.02 \times 10^{-12}$ & 0 & 8 \\
TSMixer & & & 0.06M & 0.442 & 0.0053 & 0.001 & 0.5175 & 0.9823 & 0.2013 & 0 & 1 \\
\midrule
Autoformer &
\multirow{4}{*}{ETTh1} &
\multirow{4}{*}{720} 
         & 10.54M & 0.5219 & 0.0622 & 0.0151 & 0.3825 & 0.9575 & 0.0027 & 0 & 0 \\
iTransformer & & & 1M & 0.4922 & 0.034 & 0.0053 & 0.1062 & 0.9725 & 0.0341 & 1 & 2 \\
NLinear & & & 0.07M & 0.4626 & 0.006 & 0.0008 & 3.4412 & 0.6569 & $5.94 \times 10^{-14}$ & 0 & 10 \\
TSMixer & & & 0.1M & 0.4661 & 0.0098 & 0.0021 & 0.7029 & 0.9616 & 0.0052 & 0 & 4 \\
\midrule
Autoformer & 
\multirow{4}{*}{ETTh2} &
\multirow{4}{*}{96}
         & 10.54M & 0.4136 & 0.0745 & 0.0165 & 0.7834 & 0.9444 & 0.0004 & 0 & 3 \\
iTransformer & & & 0.22M & 0.3501 & 0.0043 & 0.0009 & 0.0546 & 0.9862 & 0.3851 & 0 & 0 \\
NLinear & & & 0.01M & 0.3383 & 0.0035 & 0.0005 & 1.6811 & 0.8869 & $3.61 \times 10^{-7}$ & 0 & 4 \\
TSMixer & & & 0.04M & 0.3402 & 0.0048 & 0.0011 & 0.2689 & 0.9804 & 0.1427 & 0 & 0 \\
\midrule
Autoformer & 
\multirow{4}{*}{ETTh2} &
\multirow{4}{*}{192}
         & 10.54M & 0.4596 & 0.1138 & 0.02 & 2.1106 & 0.7849 & $8.76 \times 10^{-11}$ & 0 & 13 \\
iTransformer & & & 0.24M & 0.3989 & 0.0037 & 0.0007 & 0.5851 & 0.9712 & 0.0271 & 0 & 4 \\
NLinear & & & 0.02M & 0.3906 & 0.0161 & 0.0016 & 7.9739 & 0.3219 & $2.15 \times 10^{-19}$ & 1 & 4 \\
TSMixer & & & 0.05M & 0.3937 & 0.0052 & 0.001 & 0.5335 & 0.9691 & 0.0188 & 1 & 5 \\
\midrule
Autoformer & 
\multirow{4}{*}{ETTh2} &
\multirow{4}{*}{336}
         & 10.54M & 0.4756 & 0.0514 & 0.0091 & 1.3831 & 0.9096 & $4.10 \times 10^{-6}$ & 0 & 3 \\
iTransformer & & & 0.26M & 0.4328 & 0.007 & 0.0012 & 0.2282 & 0.9915 & 0.7858 & 1 & 1 \\
NLinear & & & 0.02M & 0.4277 & 0.0608 & 0.0062 & 8.3552 & 0.2593 & $3.58 \times 10^{-20}$ & 0 & 11 \\
TSMixer & & & 0.06M & 0.4299 & 0.012 & 0.0022 & 1.9305 & 0.8175 & $8.86 \times 10^{-10}$ & 0 & 6 \\
\midrule
Autoformer & 
\multirow{4}{*}{ETTh2} &
\multirow{4}{*}{720}
         & 10.54M & 0.4976 & 0.1318 & 0.021 & 2.5122 & 0.7663 & $2.58 \times 10^{-11}$ & 0 & 7 \\
iTransformer & & & 0.3M & 0.4475 & 0.0188 & 0.003 & 0.8107 & 0.9477 & 0.0006 & 2 & 6 \\
NLinear & & & 0.07M & 0.4448 & 0.0243 & 0.0036 & 2.6457 & 0.7535 & $1.16 \times 10^{-11}$ & 0 & 5 \\
TSMixer & & & 0.1M & 0.4548 & 0.0078 & 0.0013 & 0.9528 & 0.9399 & 0.0002 & 0 & 2 \\
\hline
\end{tabular}
} 
\caption{Performance and robustness metrics for time series forecasting models on ETTh1 and ETTh2. Each row shows results for 100 seed replicates. Columns: \# Para (parameter count), Mean (mean MAE), Range, SD (standard deviation), Skewness, Shapiro W (Shapiro-Wilk W statistic), Shapiro p (Shapiro-Wilk p-value), LFO (lower fence outliers), UFO (upper fence outliers). Models grouped by dataset and forecast horizon.}
\label{tab:ts_mae_results_1}
\end{table}
\clearpage

\clearpage
\begin{table}[h!]
\centering
\resizebox{\textwidth}{!}{%
\begin{tabular}{lcc|c|c|cc|ccc|cc}
\hline
\multirow{3}{*}{} & & & \multirow{1}{*}{} & \multirow{1}{*}{} & \multicolumn{2}{c|}{Spread} & \multicolumn{3}{c|}{Symmetry} & \multicolumn{2}{c}{Tail Risk} \\
Model & Dataset & Horizon & \# Para & Mean & Range & SD & Skewness & Shapiro W & Shapiro p & LFO & UFO \\
\midrule
Autoformer & 
\multirow{4}{*}{ETTm1} &
\multirow{4}{*}{96}
         & 10.54M & 0.4669 & 0.0802 & 0.017 & 1.1613 & 0.903 & $1.95 \times 10^{-6}$ & 0 & 4 \\
iTransformer & & & 0.22M & 0.3766 & 0.0061 & 0.0013 & 0.0208 & 0.9888 & 0.5718 & 0 & 0 \\
NLinear & & & 0.01M & 0.3759 & 0.0038 & 0.0006 & 0.9646 & 0.8816 & $2.13 \times 10^{-7}$ & 5 & 8 \\
TSMixer & & & 0.04M & 0.363 & 0.0063 & 0.0014 & 0.2746 & 0.8816 & 0.4022 & 0 & 0 \\
\midrule
Autoformer & 
\multirow{4}{*}{ETTm1} &
\multirow{4}{*}{192}
         & 10.54M & 0.4933 & 0.0537 & 0.0127 & 0.3945 & 0.9585 & 0.0032 & 0 & 0 \\
iTransformer & & & 0.24M & 0.3947 & 0.0057 & 0.0011 & 0.3378 & 0.9889 & 0.5737 & 0 & 2 \\
NLinear & & & 0.02M & 0.3951 & 0.0035 & 0.0004 & 1.7726 & 0.8534 & $1.58 \times 10^{-8}$ & 3 & 6 \\
TSMixer & & & 0.05M & 0.384 & 0.0074 & 0.0014 & 0.3375 & 0.9888 & 0.569 & 0 & 1 \\
\midrule
Autoformer & 
\multirow{4}{*}{ETTm1} &
\multirow{4}{*}{336}
         & 10.54M & 0.4899 & 0.0716 & 0.0147 & 0.0694 & 0.9914 & 0.7797 & 0 & 0 \\
iTransformer & & & 0.26M & 0.4184 & 0.005 & 0.001 & 0.2745 & 0.9828 & 0.2198 & 0 & 1 \\
NLinear & & & 0.02M & 0.4165 & 0.0024 & 0.0004 & 1.4675 & 0.8365 & $3.88 \times 10^{-9}$ & 5 & 10 \\
TSMixer & & & 0.06M & 0.4051 & 0.0085 & 0.0016 & 0.3706 & 0.9823 & 0.2017 & 0 & 0 \\
\midrule
Autoformer & 
\multirow{4}{*}{ETTm1} &
\multirow{4}{*}{720}
         & 10.54M & 0.5028 & 0.0586 & 0.0123 & 0.296 & 0.9843 & 0.2812 & 0 & 1 \\
iTransformer & & & 0.3M & 0.4582 & 0.0068 & 0.0015 & 0.4214 & 0.9825 & 0.2082 & 0 & 2 \\
NLinear & & & 0.07M & 0.451 & 0.0023 & 0.0003 & 2.0421 & 0.7568 & $1.43 \times 10^{-11}$ & 5 & 14 \\
TSMixer & & & 0.1M & 0.4412 & 0.0072 & 0.0012 & 0.409 & 0.7568 & 0.1885 & 0 & 1 \\
\midrule
Autoformer & 
\multirow{4}{*}{ETTm2} &
\multirow{4}{*}{96}
        & 10.54M & 0.3217 & 0.0999 & 0.0134 & 2.0485 & 0.864 & $4.06 \times 10^{-8}$ & 0 & 4 \\
iTransformer & & & 0.22M & 0.2701 & 0.009 & 0.002 & -0.6576 & 0.9401 & 0.0002 & 0 & 0 \\
NLinear & & & 0.01M & 0.2653 & 0.0045 & 0.0009 & 0.4984 & 0.9595 & 0.0037 & 2 & 6 \\
TSMixer & & & 0.04M & 0.2609 & 0.0084 & 0.0014 & 0.6217 & 0.9653 & 0.0097 & 1 & 3 \\
\midrule
Autoformer & 
\multirow{4}{*}{ETTm2} &
\multirow{4}{*}{192}
         & 10.54M & 0.3418 & 0.0331 & 0.007 & 0.8316 & 0.9469 & 0.0005 & 0 & 2 \\
iTransformer & & & 0.24M & 0.3124 & 0.006 & 0.0011 & 0.2712 & 0.991 & 0.7461 & 0 & 1 \\
NLinear & & & 0.02M & 0.3048 & 0.0038 & 0.0006 & 0.6495 & 0.9517 & 0.0011 & 1 & 2 \\
TSMixer & & & 0.05M & 0.3027 & 0.0075 & 0.0012 & -0.3363 & 0.9853 & 0.3335 & 1 & 0 \\
\midrule
Autoformer & 
\multirow{4}{*}{ETTm2} &
\multirow{4}{*}{336}
         & 10.54M & 0.3731 & 0.0285 & 0.0059 & 0.9005 & 0.9446 & 0.0004 & 0 & 3 \\
iTransformer & & & 0.26M & 0.3518 & 0.006 & 0.0011 & -0.2772 & 0.9885 & 0.5431 & 2 & 1 \\
NLinear & & & 0.02M & 0.343 & 0.0043 & 0.0007 & 1.1313 & 0.9132 & $6.24 \times 10^{-6}$ & 0 & 4 \\
TSMixer & & & 0.06M & 0.3404 & 0.006 & 0.0014 & 0.2572 & 0.9669 & 0.013 & 0 & 0 \\
\midrule
Autoformer & 
\multirow{4}{*}{ETTm2} &
\multirow{4}{*}{720}
        & 10.54M & 0.4186 & 0.0229 & 0.0047 & 0.8522 & 0.9476 & 0.0006 & 0 & 4 \\
iTransformer & & & 0.3M & 0.406 & 0.0082 & 0.0016 & 0.1131 & 0.9929 & 0.8801 & 0 & 0 \\
NLinear & & & 0.07M & 0.3989 & 0.0027 & 0.0005 & 0.2863 & 0.984 & 0.2665 & 0 & 1 \\
TSMixer & & & 0.1M & 0.3977 & 0.008 & 0.0016 & 0.4185 & 0.9791 & 0.1127 & 0 & 3 \\
\hline
\end{tabular}
} 
\caption{Performance and robustness metrics for time series forecasting models on ETTm1 and ETTm2. Each row shows results for 100 seed replicates. Columns: \# Para (parameter count), Mean (mean MAE), Range, SD (standard deviation), Skewness, Shapiro W (Shapiro-Wilk W statistic), Shapiro p (Shapiro-Wilk p-value), LFO (lower fence outliers), UFO (upper fence outliers). Models grouped by dataset and forecast horizon.}
\label{tab:ts_mae_results_2}
\end{table}
\clearpage

\clearpage
\begin{table}[h!]
\centering
\resizebox{\textwidth}{!}{%
\begin{tabular}{lcc|c|c|cc|ccc|cc}
\hline
\multirow{3}{*}{} & & & \multirow{1}{*}{} & \multirow{1}{*}{} & \multicolumn{2}{c|}{Spread} & \multicolumn{3}{c|}{Symmetry} & \multicolumn{2}{c}{Tail Risk} \\
Model & Dataset & Horizon & \# Para & Mean & Range & SD & Skewness & Shapiro W & Shapiro p & LFO & UFO \\
\midrule
Autoformer & 
\multirow{4}{*}{Traffic} &
\multirow{4}{*}{96}
         & 14.91M & 0.393 & 0.0724 & 0.0147 & 1.2906 & 0.8991 & $1.29 \times 10^{-6}$ & 0 & 6 \\
iTransformer & & & 6.41M & 0.2686 & 0.0025 & 0.0005 & -0.0472 & 0.9915 & 0.7809 & 0 & 0 \\
NLinear & & & 0.01M & 0.3883 & 0.0056 & 0.001 & 0.9392 & 0.9471 & 0.0005 & 0 & 2 \\
TSMixer & & & 6.27M & 0.3135 & 0.0138 & 0.0025 & 0.8202 & 0.9601 & 0.0041 & 0 & 2 \\
\midrule
Autoformer & 
\multirow{4}{*}{Traffic} &
\multirow{4}{*}{192}
         & 14.91M & 0.4073 & 0.1007 & 0.0232 & 0.5618 & 0.9542 & 0.0016 & 0 & 0 \\
iTransformer & & & 6.46M & 0.2771 & 0.0019 & 0.0004 & -0.0395 & 0.9542 & 0.8782 & 0 & 0 \\
NLinear & & & 0.02M & 0.3642 & 0.0039 & 0.0007 & 1.2948 & 0.9121 & $5.49 \times 10^{-6}$ & 0 & 5 \\
TSMixer & & & 6.28M & 0.3206 & 0.0121 & 0.0022 & 0.5268 & 0.9804 & 0.1421 & 0 & 1 \\
\midrule
Autoformer & 
\multirow{4}{*}{Traffic} &
\multirow{4}{*}{336}
         & 14.91M & 0.3923 & 0.082 & 0.0157 & 1.1026 & 0.9253 & $2.72 \times 10^{-5}$ & 0 & 4 \\
NLinear & & & 0.02M & 0.3669 & 0.0037 & 0.0006 & 1.1692 & 0.9359 & 0.0001 & 0 & 3 \\
TSMixer & & & 6.3M & 0.3269 & 0.0102 & 0.0023 & 0.0592 & 0.9859 & 0.3676 & 0 & 0 \\
\midrule
Autoformer & 
\multirow{4}{*}{Traffic} &
\multirow{4}{*}{720}
         & 14.91M & 0.4058 & 0.0507 & 0.0117 & 0.4604 & 0.9732 & 0.0388 & 0 & 0 \\
NLinear & & & 0.07M & 0.387 & 0.003 & 0.0005 & 0.82 & 0.9556 & 0.002 & 0 & 5 \\
TSMixer & & & 6.33M & 0.3459 & 0.0137 & 0.0029 & 0.5306 & 0.9733 & 0.0395 & 0 & 2 \\
\midrule
Autoformer & 
\multirow{4}{*}{Weather} &
\multirow{4}{*}{96}
         & 10.61M & 0.3345 & 0.0685 & 0.0159 & 0.3316 & 0.9777 & 0.0878 & 0 & 0 \\
iTransformer & & & 4.83M & 0.2152 & 0.0111 & 0.0021 & 0.7432 & 0.9641 & 0.008 & 0 & 3 \\
NLinear & & & 0.01M & 0.2408 & 0.0028 & 0.0005 & 0.6691 & 0.9623 & 0.0059 & 1 & 6 \\
TSMixer & & & 0.12M & 0.2039 & 0.0069 & 0.0011 & 0.009 & 0.9933 & 0.9069 & 1 & 1 \\
\midrule
Autoformer & 
\multirow{4}{*}{Weather} &
\multirow{4}{*}{192}
         & 10.61M & 0.3686 & 0.0757 & 0.0146 & 0.8685 & 0.9414 & 0.0002 & 0 & 4 \\
iTransformer & & & 4.88M & 0.2578 & 0.0067 & 0.0012 & 0.4555 & 0.9798 & 0.1293 & 0 & 1 \\
NLinear & & & 0.02M & 0.2768 & 0.0024 & 0.0004 & 0.8797 & 0.9533 & 0.0014 & 0 & 3 \\
TSMixer & & & 0.13M & 0.2491 & 0.0063 & 0.0015 & 0.2265 & 0.9684 & 0.0166 & 0 & 0 \\
\midrule
Autoformer & 
\multirow{4}{*}{Weather} &
\multirow{4}{*}{336}
         & 10.61M & 0.3971 & 0.0691 & 0.014 & 0.8766 & 0.9218 & $1.75 \times 10^{-5}$ & 0 & 1 \\
iTransformer & & & 4.96M & 0.2993 & 0.0042 & 0.0008 & 0.1191 & 0.9927 & 0.8718 & 0 & 1 \\
NLinear & & & 0.02M & 0.3129 & 0.0024 & 0.0004 & 0.8446 & 0.9597 & 0.0038 & 0 & 2 \\
TSMixer & & & 0.14M & 0.2932 & 0.0117 & 0.0024 & 0.3511 & 0.9869 & 0.4279 & 0 & 0 \\
\midrule
Autoformer & 
\multirow{4}{*}{Weather} &
\multirow{4}{*}{720}
         & 10.61M & 0.4296 & 0.0722 & 0.0107 & 1.9676 & 0.8604 & $2.94 \times 10^{-8}$ & 0 & 3 \\
iTransformer & & & 5.15M & 0.3508 & 0.0052 & 0.001 & 0.0062 & 0.9949 & 0.9728 & 0 & 0 \\
NLinear & & & 0.07M & 0.36 & 0.0022 & 0.0004 & 1.4506 & 0.8807 & $1.94 \times 10^{-7}$ & 0 & 6 \\
TSMixer & & & 0.18M & 0.3454 & 0.0094 & 0.0021 & 0.4367 & 0.9654 & 0.0099 & 0 & 0 \\
\hline
\end{tabular}
} 
\caption{Performance and robustness metrics for time series forecasting models on Traffic and Weather. Each row shows results for 100 seed replicates. Columns: \# Para (parameter count), Mean (mean MAE), Range, SD (standard deviation), Skewness, Shapiro W (Shapiro-Wilk W statistic), Shapiro p (Shapiro-Wilk p-value), LFO (lower fence outliers), UFO (upper fence outliers). Models grouped by dataset and forecast horizon. Traffic horizons 336 and 720 were not completed for iTransformer due to GPU memory constraints on the NVIDIA A100s.}
\label{tab:ts_mae_results_3}
\end{table}
\clearpage

\begin{table}[!t]
\centering
\resizebox{\textwidth}{!}{%
\footnotesize
\setlength{\tabcolsep}{4pt}
\begin{tabular}{lccccc}
\toprule
\textbf{Robustness Measure} & \textbf{W} & \textbf{p-value} & \textbf{r} & \textbf{More Robust} & \textbf{Effect Size} \\
\midrule
\multicolumn{6}{l}{\textit{Cross-architecture Wilcoxon Comparison}} \\
\midrule
Range                & 1163.0  & $\boldsymbol{1.17 \times 10^{-14}}$ & \phantom{$-$}0.758 & Smaller & Large \\
Standard Deviation   & 934.0   & $\boldsymbol{2.29 \times 10^{-16}}$ & \phantom{$-$}0.805 & Smaller & Large \\
Skewness             & 3203.0  & \textbf{0.0007}                     & $-$0.332           & Larger  & Medium \\
Upper Fence Outliers & 2505.5  & \textbf{0.0004}                     & $-$0.364           & Larger  & Medium \\
\bottomrule
\end{tabular}
} 
\caption{Wilcoxon signed-rank test results for robustness comparisons between time series forecasting models with different parameter counts. W Statistic and p-value report the Wilcoxon signed-rank test statistic and corresponding p-value for paired comparisons on raw metric values. r denotes the rank-biserial correlation coefficient. More Robust indicates which model size exhibited lower metric values on the robustness measure; --- indicates no statistically significant difference. Effect Size reports the rank-biserial correlation magnitude using the thresholds: negligible ($|r| < 0.1$), small ($0.1 \leq |r| < 0.3$), medium ($0.3 \leq |r| < 0.5$), and large ($|r| \geq 0.5$). Statistically significant results ($p < 0.05$) in bold.}
\label{tab:ts_wilcoxon}
\end{table}

\section{Additional Figures}
\label{sec:additional_figures}

\subsection{Image Classification}

\begin{figure}[H]
    \centering
    \includegraphics[width=1\linewidth]{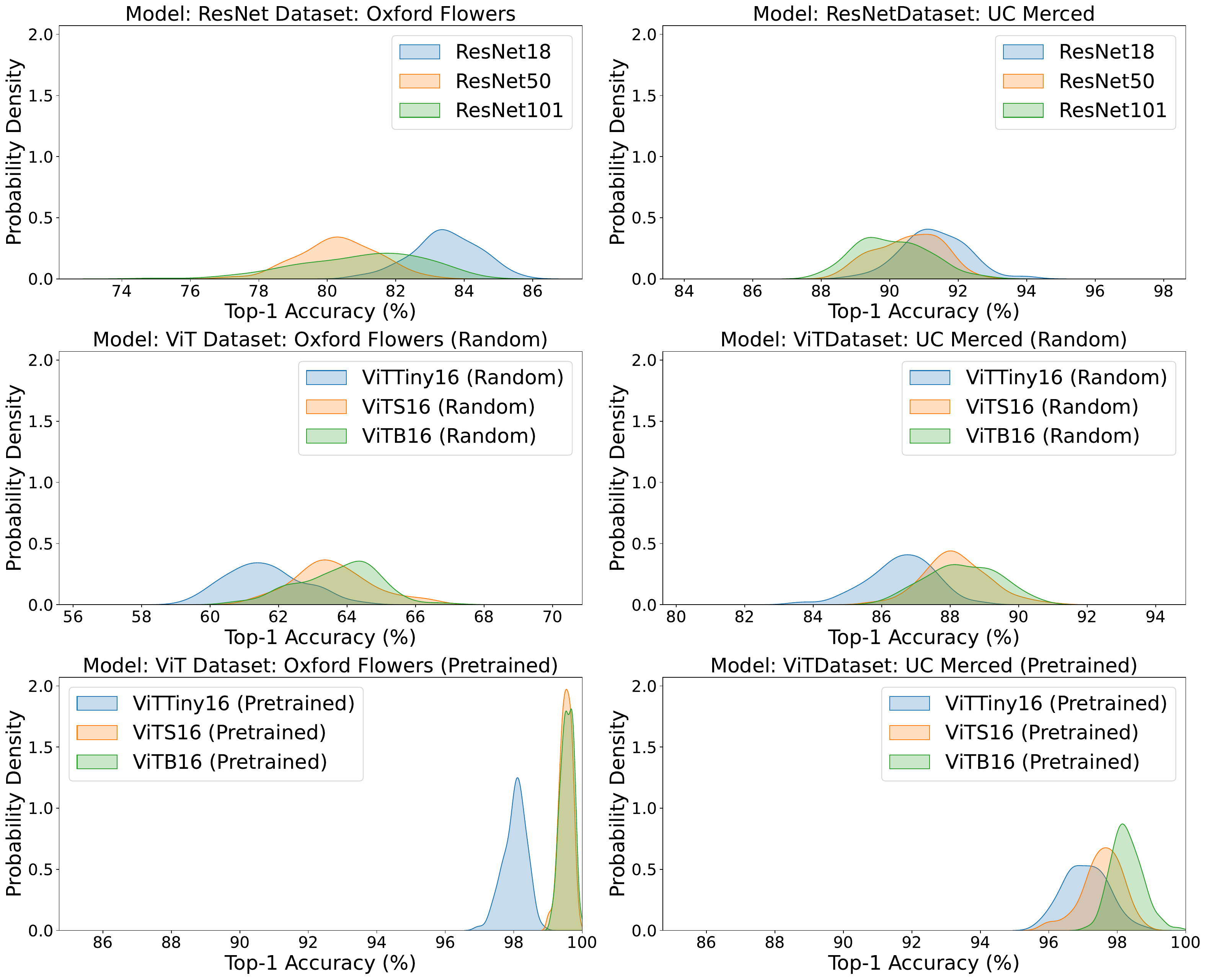}
    \caption{Kernel density estimates of top-1 accuracy for image classification models on Oxford Flowers (left) and UC Merced (right). Top row shows ResNet architectures (ResNet18, ResNet50, ResNet101); middle row shows Vision Transformers with random initialization (ViT-Tiny-16, ViT-S-16, ViT-B-16); bottom row shows Vision Transformers with pretrained weights (ViT-Tiny-16, ViT-S-16, ViT-B-16). Each curve represents the distribution across 100 random seeds for a single model-dataset pair.}
    \label{figure:results_oxucm_kde}
\end{figure}

\subsection{Time Series Forecasting}

\begin{figure}[H]
    \centering
    \includegraphics[width=1\linewidth]{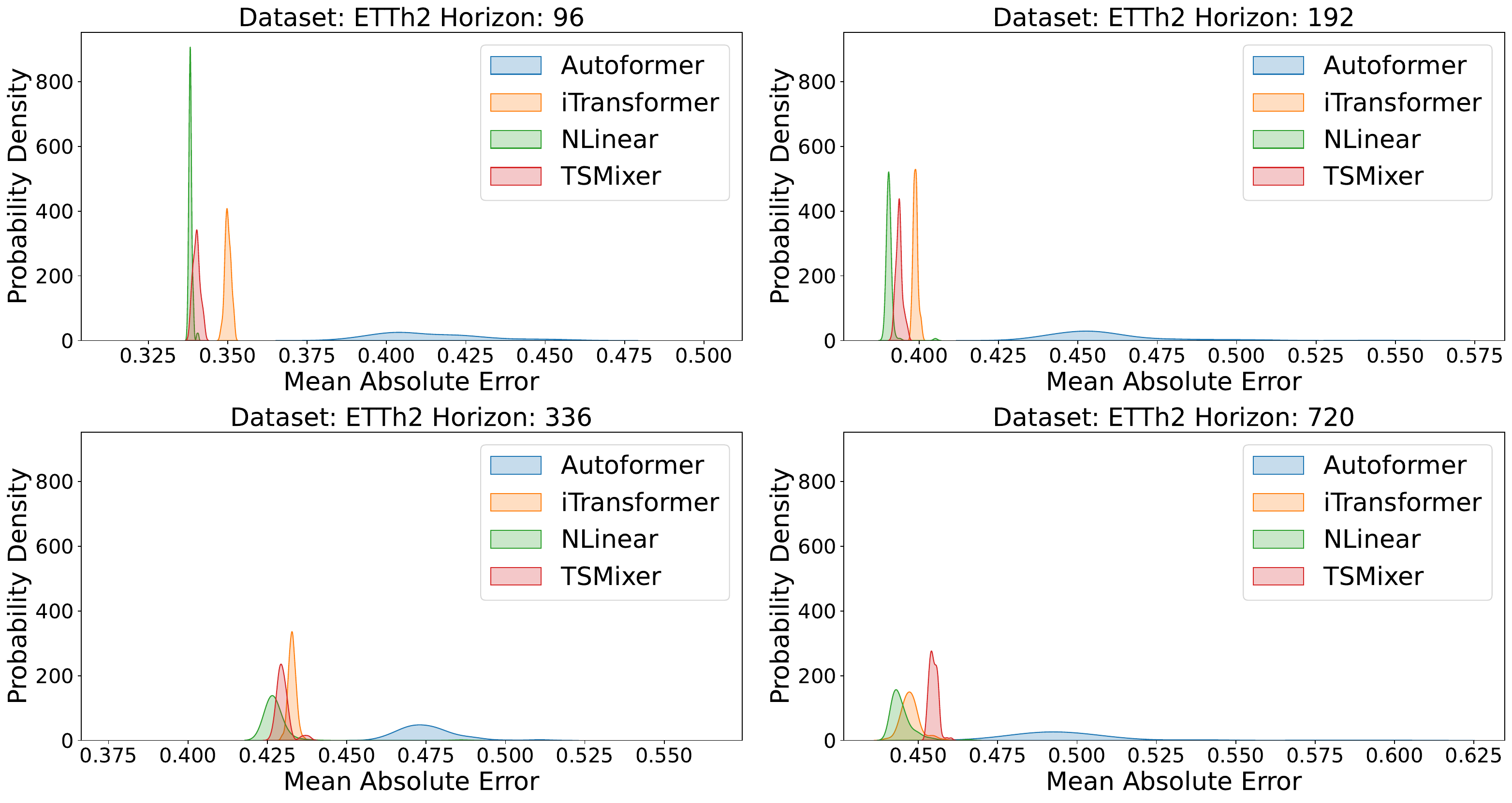}
    \caption{Kernel density estimates of MAE for time series forecasting models on ETTh2 across forecast horizons of 96, 192, 336, and 720. Each panel shows distributions for Autoformer, iTransformer, NLinear, and TSMixer. Each curve represents the distribution across 100 random seeds for a single model-horizon pair.}
    \label{figure:results_etth2_kde}
\end{figure}

\begin{figure}[H]
    \centering
    \includegraphics[width=1\linewidth]{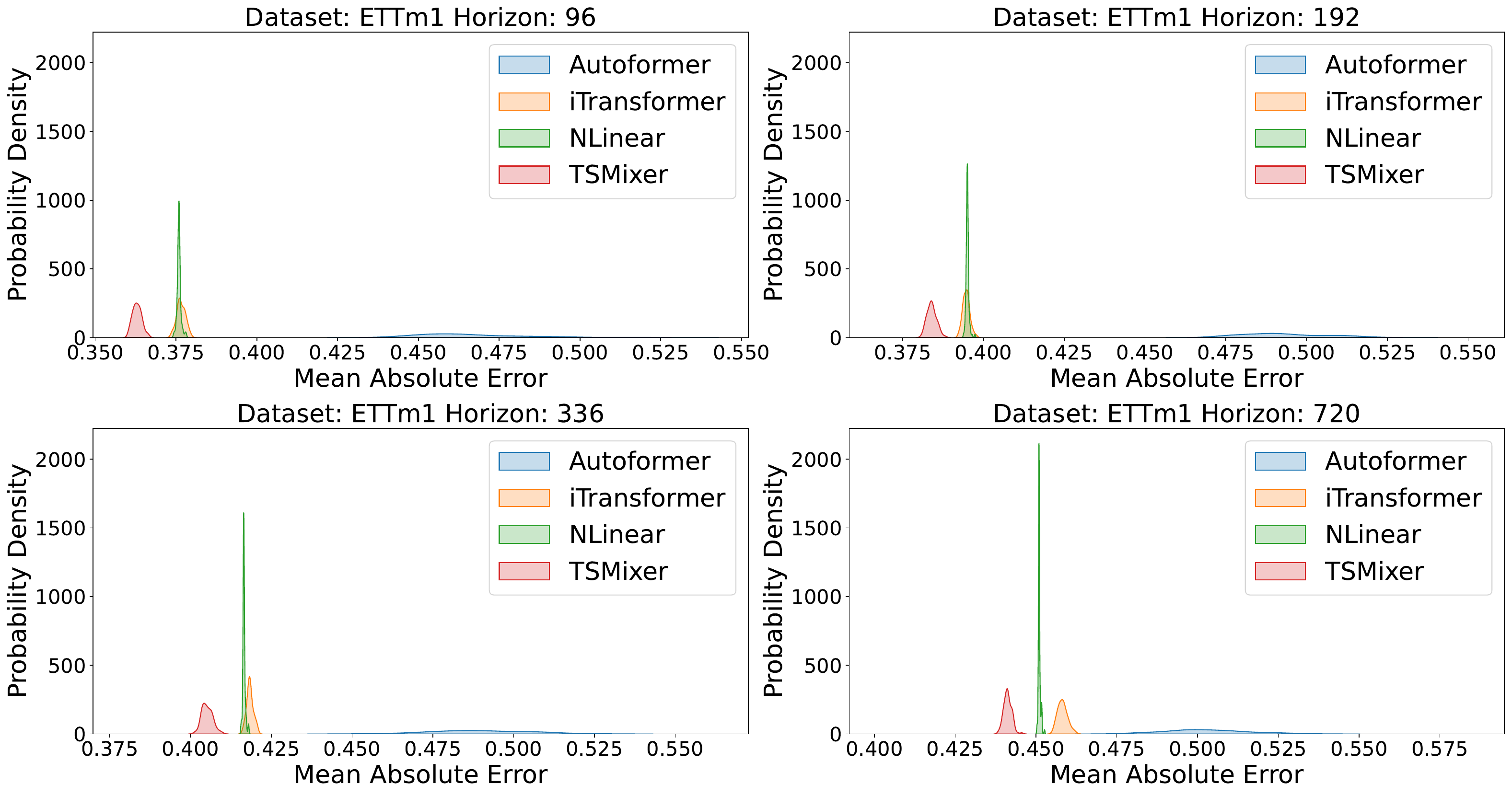}
    \caption{Kernel density estimates of MAE for time series forecasting models on ETTm1 across forecast horizons of 96, 192, 336, and 720. Each panel shows distributions for Autoformer, iTransformer, NLinear, and TSMixer. Each curve represents the distribution across 100 random seeds for a single model-horizon pair.}
    \label{figure:results_ettm1_kde}
\end{figure}

\begin{figure}[H]
    \centering
    \includegraphics[width=1\linewidth]{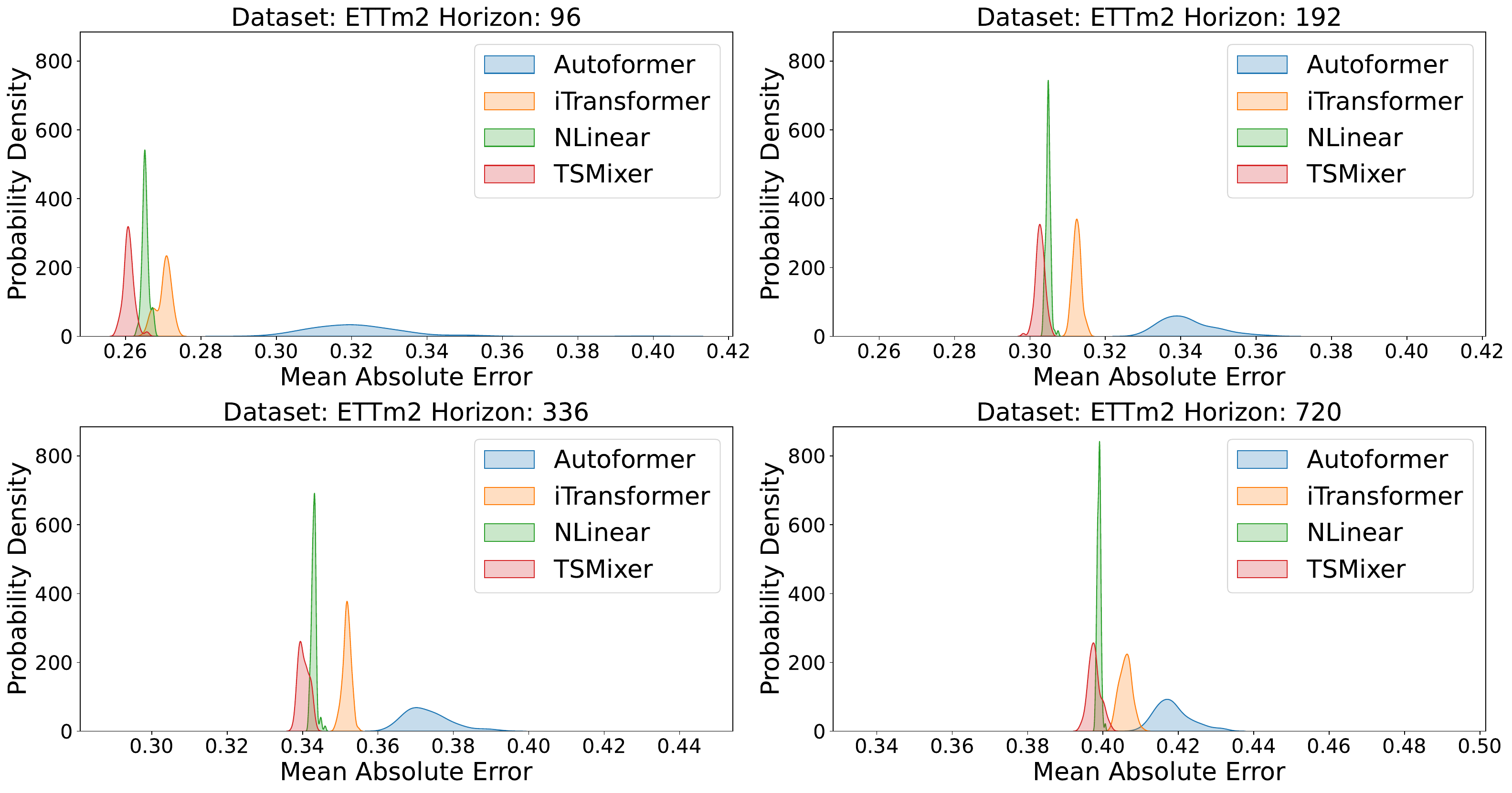}
    \caption{Kernel density estimates of MAE for time series forecasting models on ETTm2 across forecast horizons of 96, 192, 336, and 720. Each panel shows distributions for Autoformer, iTransformer, NLinear, and TSMixer. Each curve represents the distribution across 100 random seeds for a single model-horizon pair.}
    \label{figure:results_ettm2_kde}
\end{figure}

\begin{figure}[H]
    \centering
    \includegraphics[width=1\linewidth]{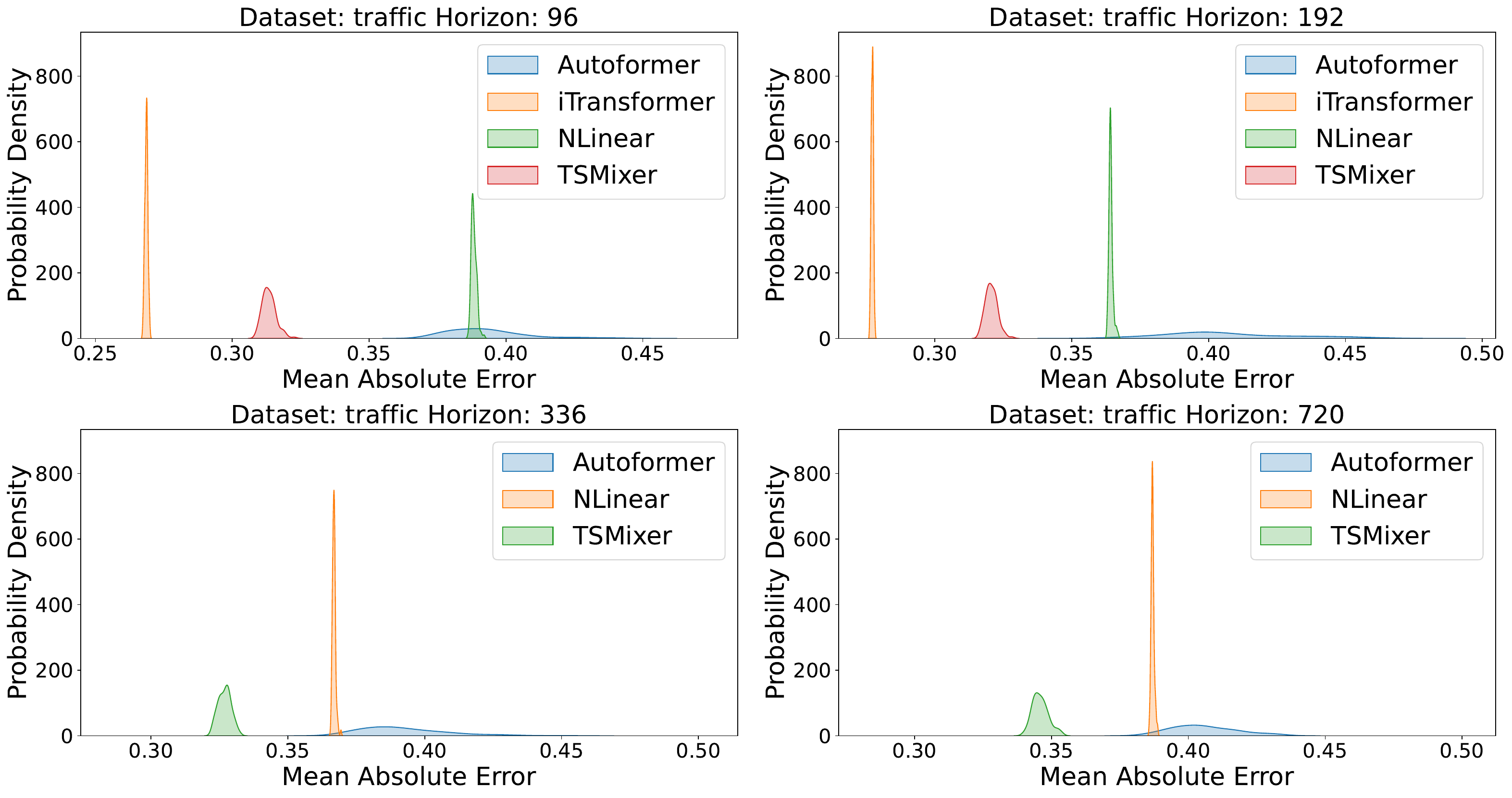}
    \caption{Kernel density estimates of MAE for time series forecasting models on Traffic across forecast horizons of 96, 192, 336, and 720. Each panel shows distributions for Autoformer, iTransformer, NLinear, and TSMixer. Each curve represents the distribution across 100 random seeds for a single model-horizon pair.}
    \label{figure:results_traffic_kde}
\end{figure}

\begin{figure}[H]
    \centering
    \includegraphics[width=1\linewidth]{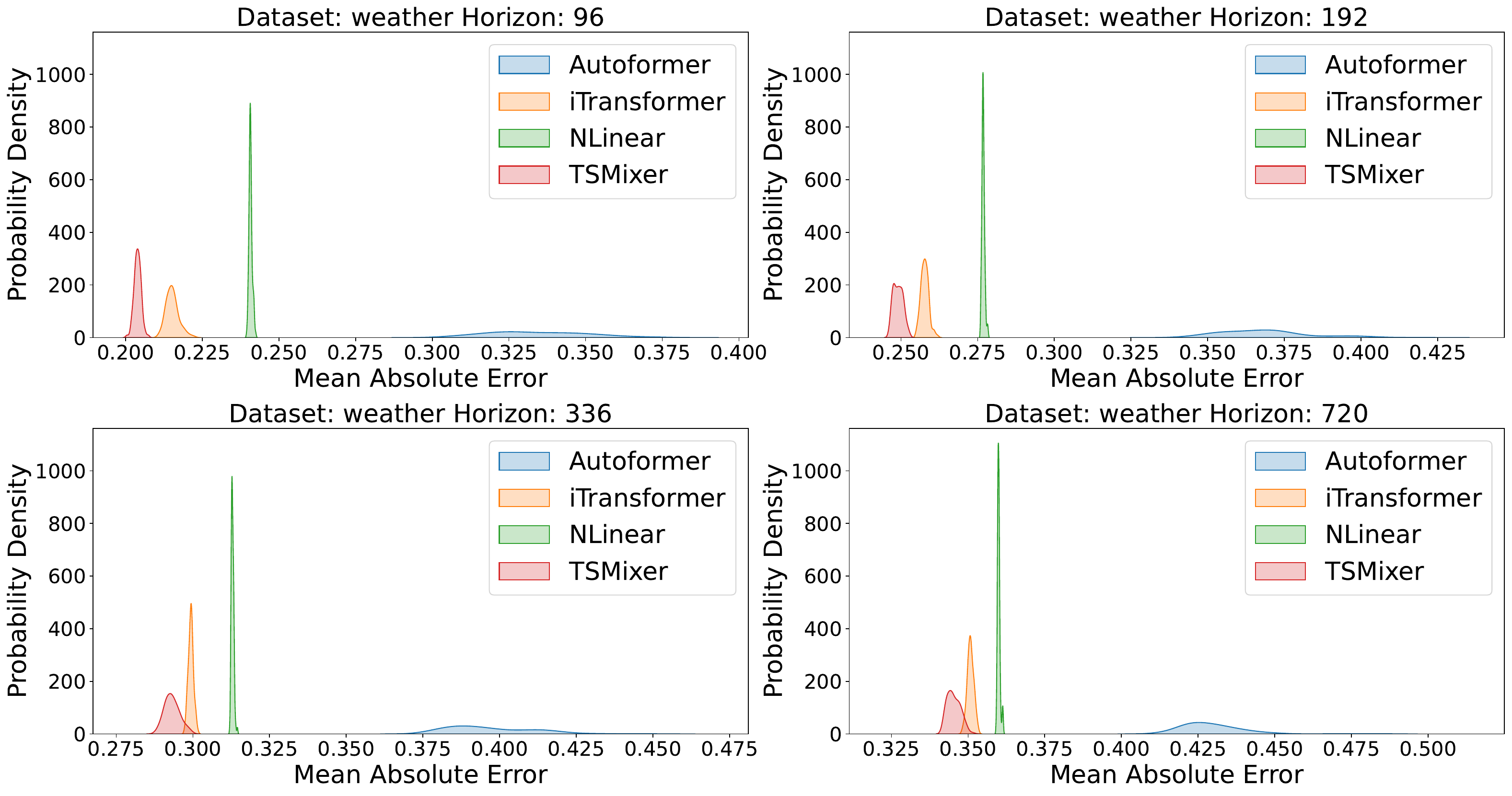}
    \caption{Kernel density estimates of MAE for time series forecasting models on Weather across forecast horizons of 96, 192, 336, and 720. Each panel shows distributions for Autoformer, iTransformer, NLinear, and TSMixer. Each curve represents the distribution across 100 random seeds for a single model-horizon pair.}
    \label{figure:results_weather_kde}
\end{figure}

\section{Gradient Dynamics and Loss Surface Curvature}

\begin{figure}[H]
    \centering
    \includegraphics[width=1\linewidth]{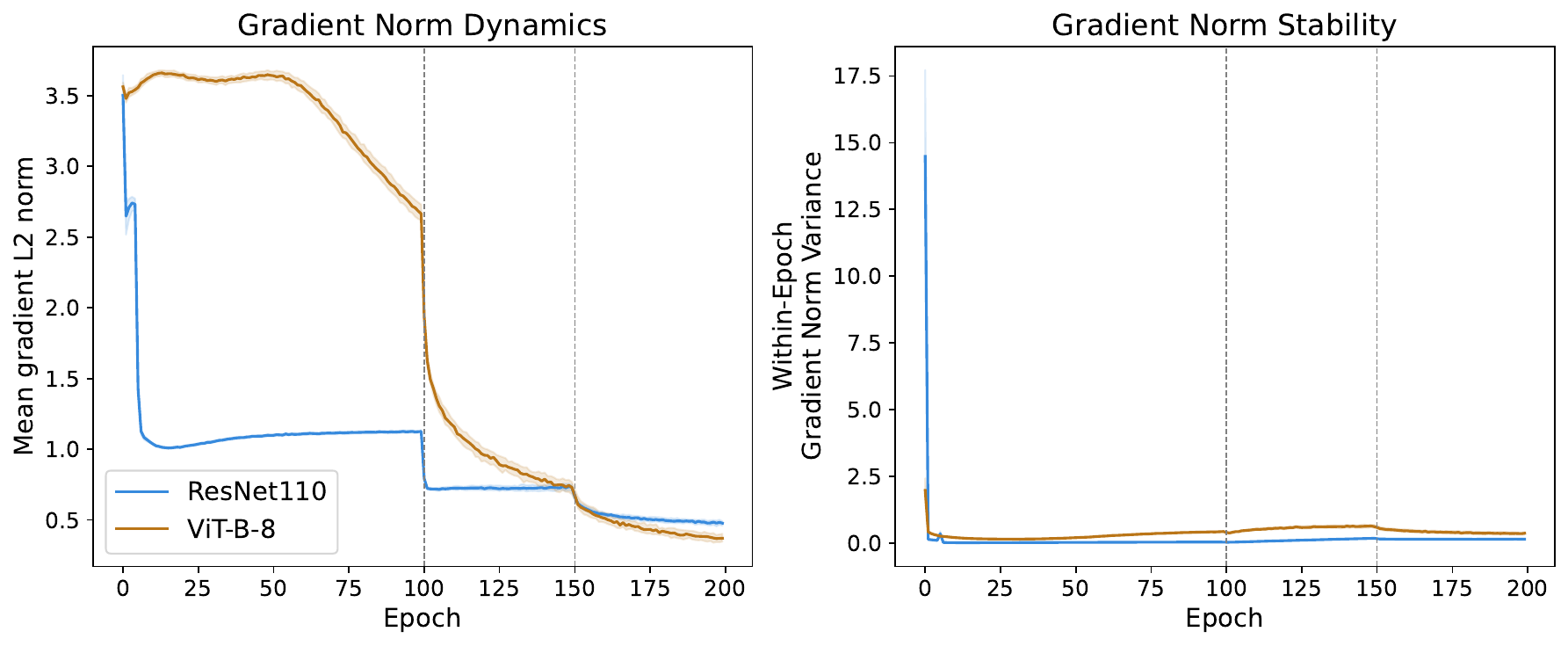}
    \caption{Mean gradient L2 norm (left) and within-epoch gradient norm variance (right) across 200 training epochs for ResNet110 and ViT-B-8 on CIFAR-10, computed over 100 seeds. Shaded bands indicate the interquartile range across seeds. Dashed vertical lines mark learning rate decay events at epochs 100 and 150.}
    \label{figure:gradient_analysis_panels_1_2}
\end{figure}

\begin{figure}[H]
    \centering
    \includegraphics[width=.5\linewidth]{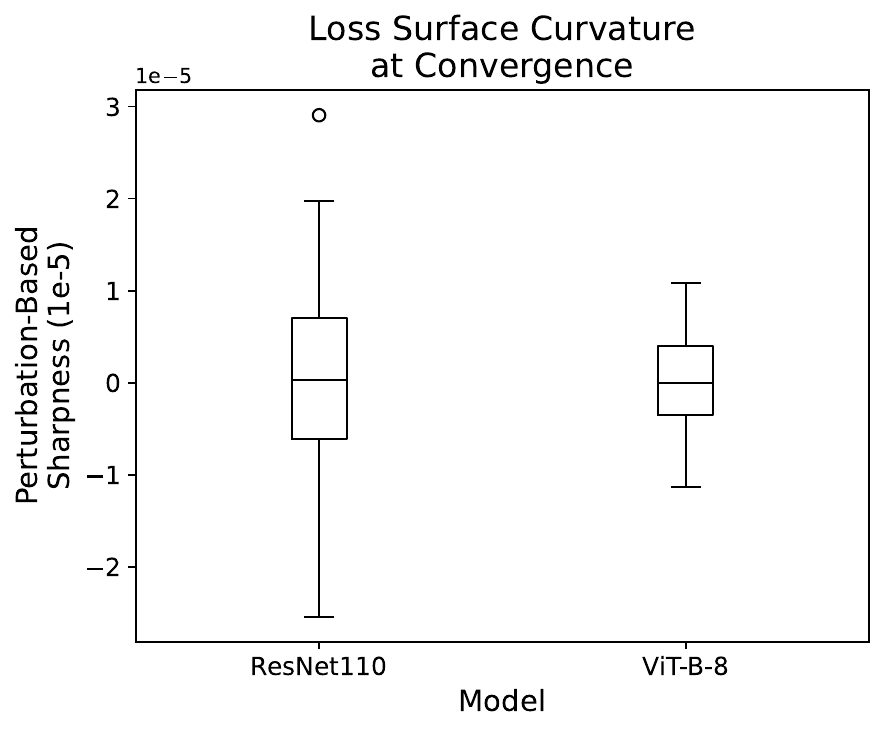}
    \caption{Perturbation-based sharpness at convergence for ResNet110 and ViT-B-8 on CIFAR-10, computed over 100 seeds. Each box shows the median, interquartile range, and 1.5 x IQR whiskers; points beyond the whiskers are plotted individually.}
    \label{figure:gradient_analysis_panel_3}
\end{figure}

We conducted an exploratory analysis of gradient norm variance and loss surface curvature for ResNet110 and ViT-B-8 on CIFAR-10, computed over 100 seeds. 
For each seed replicate, we logged the mean and within-epoch variance of the gradient L2 norm at every epoch, and computed a perturbation-based sharpness measure at the end of training following \citet{keskar_2017}. 
Sharpness was defined as the relative increase in validation loss under a random weight perturbation of norm $\epsilon = 10^{-4}$, averaged over ten perturbation samples per seed.

\autoref{figure:gradient_analysis_panels_1_2} shows the gradient dynamics across 200 training epochs. 
ViT-B-8 exhibits a mean gradient norm approximately 3$\times$ higher than ResNet110 during the first 25 epochs, after which both models converge to comparable values near 0.5. 
Within-epoch gradient norm variance collapses rapidly during the first ten epochs for both architectures; ResNet110 begins with a much larger initial spike than ViT-B-8.
The IQR bands across seeds are tight throughout, indicating that these dynamics are consistent across random initializations.

\autoref{figure:gradient_analysis_panel_3} shows the final sharpness values across 100 seeds for each model. 
Both models are centered near zero sharpness (scale $\times 10^{-5}$), with no large difference in central tendency, although ResNet110 appears to show somewhat greater variability across seeds.

These results suggest that gradient norm variance and loss surface curvature do not, by themselves, explain the differences in performance distribution spread between these architectures reported in \autoref{tab:ic_results}. 
The robustness differences observed between ResNet110 and ViT-B-8 are not clearly matched by differences in late-training optimization dynamics, indicating that the distributional differences may instead reflect architecture-specific sensitivity to initialization or early training trajectory.

\end{appendices}

\bibliography{bibliography}


\end{document}